\documentclass[
journal=medium,
manuscript=article-type,
year=2023
]{cup-journal}

\usepackage[hidelinks]{hyperref}
\usepackage{amsmath}
\usepackage[nopatch]{microtype}
\usepackage{booktabs}
\usepackage{graphicx}
\usepackage{multirow}
\usepackage{adjustbox}
\usepackage{array}
\usepackage{float}
\usepackage{placeins}
\usepackage{longtable}

\usepackage[
textwidth=1.2cm 
, tickmarkheight=0.2cm
, textsize=footnotesize 
, disable
]{todonotes}

\title{PopBERT. Detecting populism and its host ideologies in the German Bundestag} 
\author{L. Erhard}
\email[L. Erhard]{lukas.erhard@sowi.uni-stuttgart.de}
\author{S. Hanke}
\author{U. Remer}
\affiliation{Institute for Social Sciences, University of Stuttgart, BW, Germany}
\author{A. Falenska}
\affiliation{Institute for Natural Language Processing, University of Stuttgart, BW, Germany}
\author{R. Heiberger}
\affiliation{Institute for Social Sciences, University of Stuttgart, BW, Germany}

\date{\today}


\begin{document}

\begin{abstract}
The rise of populism concerns many political scientists and practitioners, yet, the detection of its underlying language remains fragmentary.
This paper aims to provide a reliable, valid, and scalable approach to measure populist stances.
For that purpose, we created an annotated dataset based on parliamentary speeches of the German Bundestag (2013--2021).
Following the ideational definition of populism, we label moralizing references to ``the virtuous people'' or ``the corrupt elite'' as core dimensions of populist language.
To identify, in addition, how the thin ideology of populism is ``thickened'', we annotate how populist statements are attached to left-wing or right-wing host ideologies.
We then train a transformer-based model (PopBERT) as a multilabel classifier to detect and quantify each dimension.
A battery of validation checks reveals that the model has a strong predictive accuracy, provides high qualitative face validity, matches party rankings of expert surveys, and detects out-of-sample text snippets correctly.
PopBERT enables dynamic analyses of how German-speaking politicians and parties use populist language as a strategic device.
Furthermore, the annotator-level data may also be applied in cross-domain applications or to develop related classifiers.
\end{abstract}

\newpage
\listoftodos
\newpage

\section{Introduction}

Studying the rise of populism represents a particularly important case of contemporary political analysis \parencite{muddeStudyingPopulismComparative2018, hungerWhatBuzzwordSystematic2022}.
Populist parties and actors have been identified as one of the main threats to democratic values \parencite{rummensPopulismThreatLiberal2017} and drivers of increased ideological polarization \parencite{robertsPopulismPolarizationComparative2022}.
The common core of populists across the globe is their emphasis on an antagonistic relationship between \textit{the virtuous people} and \textit{the corrupt elite} \parencite{bonikowskiPopulistStyleAmerican2016, muddePopulistZeitgeist2004}.
Understanding populists' rhetorics and claims that support this antagonism is thereby key to explaining their electoral success \parencite{devreesePopulismExpressionPolitical2018}.
By focusing on populist language, researchers can develop more granular measurements regarding its dynamics and contexts in comparison to, for instance, classic survey items \parencite{klamm-etal-2023-kind}.

Yet, most efforts to detect populism in German texts do not unleash the full potential of current state-of-the-art NLP models.
Even recent (quantitative) approaches apply dictionaries to identify populist statements \parencite{grundl_populist_2022}.
Since dictionaries are simple frequency counts based on manually curated lists of phrases, they can, for instance, neither consider the context in which a word is used nor identify more abstract meanings (e.g., idioms).
Those (and other) shortcomings have been mostly overcome by transformer-based models \parencite{vaswaniAttentionAllYou2017}.
Deep learning representations of language can capture words' contextual information and long-distance dependencies so that their increased prediction accuracy sets new standards for model architectures in NLP.
In particular, the BERT model \parencite{devlin:BERTPretrainingDeep.2019} provides a pre-trained language representation that can be fine-tuned with relatively small samples and outperforms traditional ML approaches considerably \parencite{wankmullerIntroductionNeuralTransfer2022}.
Those promises notwithstanding, only a few studies have made use of transformer-based models to detect populism---one focusing on the U.S. \parencite{bonikowskiPoliticsUsualMeasuring2022}, the other dedicated to extracting entities \parencite{klamm-etal-2023-kind}.

This paper aims to extend this rather short list.
We emphasize two main contributions:\footnote{For guidance on and links to all accompanying models and data available, see this article's data availability statement.}
First, we introduce a unique dataset annotated by five specialized coders based on German plenary debates from 2013 to 2021 (8,795 sentences).
We focus on two concepts that make up ``populism as content'' \parencite{devreesePopulismExpressionPolitical2018}: \textit{anti-elitism} and \textit{people-centrism}.
To investigate how the ``thin-centered ideology'' \parencite{muddeStudyingPopulismComparative2018, muddePopulistZeitgeist2004} of populism is ``thickened'', we further annotate how it is attached to left-wing (i.e., socialism) or right-wing (i.e., nativism) host ideologies \parencite{hungerWhatBuzzwordSystematic2022}. 
Secondly, this paper presents \textit{PopBERT}, a readily applicable classifier to detect populism and its host ideology for the German language based on the $GBERT_{Large}$ model \parencite{chanGermanNextLanguage2020}.
However, as a complex phenomenon that eludes clear-cut ``right'' or ``wrong'' distinctions, populism presents a challenging task for NLP models that are most often trained on objective, binary ground-truth data \parencite{plankProblemHumanLabel2022}.
We will therefore discuss several validity checks to explain how PopBERT performs substantively to underline its capability to detect German populist language in a fine-grained, context-sensitive manner.

\section{Detecting populism in texts}

Manual content analysis has a long tradition in the social sciences, and also for the measurement of populism \parencite[for an overview, see][]{aslanidis:Measuringpopulistdiscourse.2018}.
In their pioneering work, \textcite{jagersPopulismPoliticalCommunication2006} used manual coding to measure populism in six Belgian parties' political television broadcasts.
They define populism as a communicative appeal to the people, including references to the sovereignty of the people and the popular will, which \textit{might} be complemented by anti-elitism and exclusionism of certain groups.
In contrast, \textcite{ernstBipolarPopulismUse2017} use an ideational definition of populism introduced by \textcite{muddePopulistZeitgeist2004}, in which anti-elitism plays a constitutive part to identify populist key messages in Facebook and Twitter posts.
Most recently, \textcite{schurmannYellingSidelinesHow2022} manually examined 3,500 Facebook posts and demonstrated that outsider parties more frequently combine populist and crisis-related content compared to established parties.

As every element of the corpus is individually evaluated against the codebook, manual coding may be seen as gold standard regarding the validity of text analysis.
However, manual content analysis faces severe constraints concerning the feasible amount of data and metrics' reliability.
This forces researchers, for instance, to select samples \parencite{schurmannYellingSidelinesHow2022} or make qualitative choices on which statements should be coded \parencite{ernstBipolarPopulismUse2017}.
Therefore, reliable and scalable strategies are needed to analyze larger corpora. 

Expert-curated dictionaries represent such an approach by providing static, discrete definitions for individual words; for instance, the word ``good'' may be assigned to a positive sentiment.
Among the first to implement dictionary-based approaches to measure populism are \textcite{rooduijn:MeasuringPopulismComparing.2011}.
Following the definition of \textcite{muddePopulistZeitgeist2004}, they conceptualize populism as a thin ideology with the central concepts of people-centrism and anti-elitism.
They measure it in the manifestos of 24 parties from four Western European countries. 
The authors argue that their dictionary approach addresses weaknesses of previous studies, such as the low reliability of holistic grading \parencite{hawkins:ChavezPopulistMeasuring.2009} or the lack of addressing validity as in \textcite{jagersPopulismPoliticalCommunication2006}.
\textcite{bonikowskiPopulistStyleAmerican2016} seek to detect populist claims in U.S. presidential elections campaign speeches measured with a dictionary applied to 2,406 campaign speeches of candidates between 1952 and 1996.
Similarly, \textcite{grundl_populist_2022} develops a dictionary for measuring populism in social media postings of German-speaking politicians and parties.
Aggregated at the level of parties, Gründl's approach yields a closer approximation of expert ratings than previous dictionaries.

Although dictionary-based measurements allow scaling the amount of processed text with perfect reliability, they have serious constraints regarding the significant manual work that is needed to design rule-based heuristics accounting for even basic linguistic patterns like negations (``not good''), and, much more so, context-sensitive meanings (``good'' as adjective vs. in economics) or paraphrasing idioms (``on cloud nine'').
Furthermore, dictionaries attempting to capture populism vary widely in the number and conjugations of terms used, with no clear criterion for what an extensive dictionary might be \parencite{pauwelsMeasuringPopulismReview2017}.

These (and other) shortcomings have been tackled by next-generation language representations, in particular, by \textit{transformer} models.
The landmark model was developed by \textcite{vaswaniAttentionAllYou2017}.
The key innovation of the transformer architecture is the ability to represent words together with their contexts, i.e., different occurrences of the same word are represented by different vectors. 
Transformers achieve this by employing \textit{self-attention}, a mechanism that enables the model to attend to different parts of the input and to determine their relative importance.
Since the attention mechanism can consider the whole input sequence, it can capture long-range dependencies and effectively model complex linguistic phenomena like polysemy (e.g., multiple meanings of ``good'') or idioms (e.g., ``on cloud nine'').

However, we only identified two studies using transformer models to detect populism in texts, and another paper that uses supervised machine learning.\footnote{Studies that define populist language by the party family of the speaker are excluded. As \textcite{jankowskiWhenCorrelationNot2023} lay out in their detailed critique on \textcite{dicoccoHowPopulistAre2022}, deriving a measure by learning a classifier based on the differences in the communicative styles of groups is insufficient. The classification then just resembles the measurement of the group-discerning feature plus some error due to misclassification.}
\textcite{bonikowskiPoliticsUsualMeasuring2022} follow \textcite{muddePopulistZeitgeist2004} and characterize populism by anti-elitism and people-centrism when communicated in a moralizing way.
They identify populism in speeches of presidential election campaigns in the U.S. at the level of paragraphs by fine-tuning six independent binary transformer-based RoBERTa models and report an average precision of .68 and recall of .61.
While \textcite{daiWhenPoliticiansUse2022} apply a similar ideational framework and use the same data source as \textcite{bonikowskiPoliticsUsualMeasuring2022}, they rely on a supervised machine learning approach without transformer-based embeddings. 
Instead, they use a binary random forest classifier with Doc2Vec embeddings as features.
Only recently, \textcite{klamm-etal-2023-kind} introduced an annotated dataset of German parliamentary speeches from 2017 to 2021 for a transformer-based model.
They label references to \textit{the people} and \textit{the elite} in a hierarchical three-level coding scheme, building on Mudde's ideational framework outlined by \textcite{wirthAppealPopulistIdeas2016}. 
While their data and theoretical background are close to our approach, they do not classify anti-elitism or people-centrism in text snippets but focus on detecting entities (e.g., organizations or persons) that refer to the people or the elite \parencite[or their antidotes;][]{klamm-etal-2023-kind}. 

Summarizing the state of the art in detecting populism in texts, we find, firstly, that most scholars apply an ideational approach, with an emphasis on anti-elitism \parencite{muddePopulistZeitgeist2004}---albeit papers differ in the breadth and depth of annotating populist statements.
Secondly, prior studies use varying lengths of coding units.
As the type of text snippets varies (sometimes even within studies through different lengths of paragraphs), this reduces comparability \parencite{aslanidis:Measuringpopulistdiscourse.2018}. 
Thirdly, few attempts to detect populist language exist that exploit the possibilities of state-of-the-art NLP models.
To utilize the advantages of transformer-based classifications, we need to elaborate on the theoretical understanding of populism first in order to derive concise annotations.

\section{Conceptualizing Populism}\label{sec:theory}

Populism, even in its narrow use as a scientific concept, covers different understandings of the phenomenon \parencite{rovirakaltwasserOxfordHandbookPopulism2017, hungerWhatBuzzwordSystematic2022}.
Nevertheless, the literature increasingly converges on the ideational definition \parencite{muddePopulistZeitgeist2004, hawkinsIdeationalApproachPopulism2019, bonikowskiPoliticsUsualMeasuring2022, klamm-etal-2023-kind}.
In this conceptualization, populism concerns an antagonistic relationship between the corrupt elite and the virtuous people.
It thus rests on two central concepts: \textit{anti-elitism} and \textit{people-centrism}.
While populism is viewed as a ``thin-centered ideology'' \parencite{muddeStudyingPopulismComparative2018, muddePopulistZeitgeist2004}---a set of ideas with limited programmatic scope compared to ``thick ideologies'' like nativism or socialism (to which populism may be attached)---, it is often used in discourses to frame content \parencite{devreesePopulismExpressionPolitical2018}.
 
Populists claim to express the ``will of the people'' \parencite{hawkinsIdeationalApproachPopulism2019}, in which the people are characterized as homogenous or monolithic \parencite{marchRadicalLeftParties2012,albertazziTwentyfirstCenturyPopulism2008,jagersPopulismPoliticalCommunication2006}.
However, who belongs to these people is only an ``imagined heartland'' \parencite{muddePopulistZeitgeist2004,taggartPopulism2000}.
Conceptions of the people may be, for instance, cultural (as a nation or ethnos) or economic \parencite[as a class-based understanding of deprived citizens;][]{kriesiPopulistChallenge2014,canovanTakingPoliticsPeople2002}.
Similarly to the differing interpretations of the people, the elite depends on the context and host ideology.
Different groups may be targeted, such as political, economic, cultural, intellectual, or legal elites \parencite{albertazziTwentyfirstCenturyPopulism2008,jagersPopulismPoliticalCommunication2006}.

If the attached host ideology of populists is nativism, scholars speak of \textit{right-wing populism} \parencite{muddePopulistRadicalRight2007}.
Nativism is defined as `an ideology, which holds that states should be inhabited exclusively by members of the native group (``the nation'') and that non-native elements (persons and ideas) are fundamentally threatening to the homogenous nation-state' \parencite[][18]{muddePopulistRadicalRight2007}.
Essentially, nativism is an exclusionary worldview that primarily defines the people by who and what does \textit{not} belong to it: immigrants, asylum seekers or (indigenous) ethnic minorities, ``sexual deviants'', feminists, welfare recipients, or international organizations \parencite{muddeExclusionaryVsInclusionary2013, rooduijnPopulistZeitgeistProgrammatic2014}.

In contrast, \textit{left-wing populism} claims to represent the interests of marginalized groups such as immigrants, LGBTQ, women, disabled or unemployed people \parencite{muddeExclusionaryVsInclusionary2013, marchLeftRightPopulism2017}.
Thus, left-wing populism focuses on \textit{class} as its defining feature.
It frames the people mainly as ``the deprived'' in socio-economic terms and elites as part of profiteers of the capitalist (neo-liberal, market-oriented) system \parencite{marchRadicalLeftParties2012}.

Regardless of left-wing or right-wing ideologies, the relationship between people and elites is described as oppositional and antagonistic, representing a \textit{moral dichotomy} \parencite{muddePopulistZeitgeist2004, hawkins:ChavezPopulistMeasuring.2009, hawkinsIdeationalApproachPopulism2019a, daiWhenPoliticiansUse2022}. 
This Manichean struggle defines \textit{virtuous} people who oppose a \textit{corrupt} elite \parencite{hawkins:ChavezPopulistMeasuring.2009, rovirakaltwasserOxfordHandbookPopulism2017}. 
Populism is thus based on a moral divide: the people are understood as good and pure, and the elite as bad and evil.
Hence, the people are the only legitimate source of power \parencite{aslanidis:Measuringpopulistdiscourse.2018}, while the elite deprives the people of rights, values, wealth, or even their identity \parencite{albertazziTwentyfirstCenturyPopulism2008, aslanidis:Measuringpopulistdiscourse.2018}. 

Against the view that moralistic language is a constituting dimension of populism, \textcite{stavrakakis:Accomplishmentslimitationsnew.2018} argue that morality is ill-defined and too broad to constitute a necessary feature.
Yet, \textcite{stavrakakis:Accomplishmentslimitationsnew.2018} neglect that moralistic language in itself may not be a sufficient criterion to determine whether a statement is populist, but that it becomes populist only when it is \textit{combined} with anti-elitist and people-centric statements.
Like other recent studies, we therefore consider a moralizing frame as a third condition for any populist statement and take it as a modeling challenge to measure it \parencite{daiWhenPoliticiansUse2022, bonikowskiPoliticsUsualMeasuring2022, mudde:PopulismIdeationalApproach.2017, klamm-etal-2023-kind}.
Thus, we view all three defining characteristics as necessary conditions of populist language.

\section{Data \& Methods}\label{sec:methods}

\subsection{Model architecture}\label{sec:corpus}

If key components of the populist worldview (i.e., anti-elitism, people-centrism, each presented in a moralistic tone) are expressed publicly, it is possible to determine the relative strength of populist language rather than categorizing actors as populist or non-populist a priori \parencite{aslanidisPopulismIdeologyRefutation2016,hawkinsIdeationalApproachPopulism2019}. 
This means that instead of understanding an actor as populist or non-populist, we qualify their communication as more or less populist \parencite{rooduijnPopulistZeitgeistProgrammatic2014, devreesePopulismExpressionPolitical2018}.

For that purpose, we rely on fine-tuning a BERT model.
BERT (Bidirectional Encoder Representations from Transformers) is the most well-known transformer-based pre-trained model for English \parencite{devlin:BERTPretrainingDeep.2019}. 
While the seminal paper by \textcite{vaswaniAttentionAllYou2017} used a left-to-right architecture in which every token only attended to previous ones, \textcite{devlin:BERTPretrainingDeep.2019} incorporated context from both directions. 
The procedure involves masking a random subset of words in a given input and training the model to predict the missing words based on their context. 
Thus, while the essence of BERT is relatively simple---predict each of the masked words by its context---pre-training the model on large-scale corpora provides enough contextualized information to capture a deep understanding of language.

In particular, we rely on $GBERT_{large}$ \parencite{chanGermanNextLanguage2020}.\footnote{$GBERT_{Large}$ uses 335 million parameters compared to 110 million in $BERT_{Base}$ and clearly outperforms its smaller variant \parencite{chanGermanNextLanguage2020}.}
We utilize the pooled output of the trained model, which is the last hidden state of the CLS representation.
A linear layer and a tanh activation function follow.
Building upon this, we employ a fully connected layer with a sigmoid activation function and choose binary cross-entropy loss as our loss function.
By doing so, we follow common settings for this model and task.\footnote{In tasks like these, the concern is sometimes how to handle text segments that exceed the maximum supported length of the model. In our case, this limit is no issue using $GBERT_{Large}$ with an input size of 1024 tokens and performing sentence-level classification.}

\subsection{Corpus}\label{sec:corpus}

We fine-tune $GBERT_{Large}$ with parliamentary debates of the German Bundestag.
Fine-tuning leads to improved downstream performance and reduces the need for heavy engineering of task-specific architectures \parencite{wankmullerIntroductionNeuralTransfer2022}.
Moreover, it usually requires fewer training instances than a standard supervised model that would learn the task ``from scratch'', especially in the case of imbalanced data \parencite{millerActiveLearningApproaches2020}.

We collect the parliamentary debates from Open Discourse, which provides a relational database with full-text data of all plenary Bundestag speeches since 1949 \parencite{richterOpenDiscourse2020}.
We define all verbal acts delivered by an actor during a single session of the Bundestag as a speech.
However, we only use speeches assigned to a political faction, leaving us with a final corpus of 32,348 speeches.\footnote{A tabular listing of speeches by party and term can be found in \ref{ap_dataset}.}
To ensure comparability of language comprehension within the data set, we focus on a period covering the 18th and 19th legislative terms (October 2013 -- October 2021).
We chose this time frame because the AfD (``Alternative für Deutschland'', Alternative for Germany) entered the parliament in 2017. 
Thus, our observation time spans their appearance and a prior comparison period.

One of the practical challenges in detecting populism texts is determining the appropriate length of text snippets \parencite{klamm-etal-2023-kind}.
The duration of contributions in the German Bundestag also varies significantly, since speeches are delivered orally and transcribed by parliamentary staff so that segmentations (e.g., paragraphs) are often neither reliable nor comparable \parencite{aslanidis:Measuringpopulistdiscourse.2018}.
Therefore, our investigation uses the fine-grained level of grammatical sentences, allowing researchers to aggregate scores on the desired level (e.g., speeches, speakers, or parties).
Furthermore, due to the structure of speech in the Bundestag, we exclude the initial sentence from each consecutive speech contribution.
This exclusion is based on the observation that these initial sentences predominantly consist of greeting or response phrases, which we aim to avoid in our sample.

\subsection{Annotation process}

The performance of the whole classification task rests, to large parts, on the quality of the annotations that underlie the fine-tuning of BERT.
However, the literature does not provide universally valid recommendations on annotating, particularly when it comes to fuzzy and ambiguous concepts like populism that elude a strict definition \parencite{umaLearningDisagreementSurvey2021}.
To detect populism in texts, we rely on specialized annotators instead of crowd workers.\footnote{The coders are five political science majors (three Master's, two Bachelor's). They underwent extensive training before and during the multi-stage annotation process to clarify the dimensions according to the codebook. The total duration of the coding process was 5.5 months.} 
As a multi-faceted concept, we created a detailed guiding codebook (see \ref{sec:codebook}) based on \textcite{wirthAppealPopulistIdeas2016, ernstBipolarPopulismUse2017} that sticks as closely as possible to the theoretical arguments outlined in Section \ref{sec:theory}.
Following theory, we identified two core dimensions, \textit{anti-elitism} and \textit{people-centrism}.
As a third condition, annotators were instructed to label these dimensions only when the sentence contained moralistic language.
Since both dimensions may appear separately, annotators assess both labels independently and assign multiple annotations per sentence if necessary (hence the need for \textit{multilabel} models outlined below).

Furthermore, annotators also coded \textit{left-wing} or \textit{right-wing} host ideology, yet only if it \textit{co-occurred} with one of the core dimensions.
For instance, if the people are described as excluding non-native groups and ideas, the text should be marked as right-wing.
When the people are viewed from a class-based perspective instead (e.g., ``the workers''), the sentence should be labeled as \textit{left-wing}.

We started the annotation process with a stratified random sample of 2858 sentences.
Therein, a first stratum ensures sufficient representation of all parties by sampling uniformly from the speeches of all parties.
Since populism is a rare event in parliamentary speeches, we used the dictionary from \textcite{grundl_populist_2022} as a second stratum.
It allowed us to draw an equal number of positive and negative scoring sentences for each party.

We ran seven additional rounds of annotations, each consisting of between 500 and 1000 sentences.
We utilized active learning to derive samples that maximize the classifiers' performance. 
Active learning is a common method in ML to reduce the labeling effort by selecting cases from which the supervised models profit most \parencite{millerActiveLearningApproaches2020}.
In our case, we focused on ambiguous cases (i.e., edge cases) and cases from underrepresented categories.

In total, this iterative process yielded 8,795 annotated sentences. 
The number of sentences by dimension can be found in column $N$ of Table \ref{table_descriptives}.
We also provide Fleiss' $\kappa$ and the percentage agreement among coders for sentences per dimension.
While the percentage agreement naturally increases as the dimension becomes less prevalent in the dataset, Fleiss' $\kappa$ represents only moderate values.
However, the agreement of our coders is comparable to other studies that also annotate rather subjective concepts like morality \parencite[e.g., ][]{kobbe-etal-2020-exploring}, emotions \parencite[e.g., ][]{wood-etal-2018-comparison} or hate speech \parencite[e.g., ][]{sapAnnotatorsAttitudesHow2022}.

\begin{table}[hbt!]
	\begin{threeparttable}
		\caption{Number of annotated sentences in the dataset. In total, 8795 sentences were annotated by five coders each. Column N indicates how many of these sentences were labeled by at least one coder with the respective dimension. The remaining two columns provide information about the inter-annotator agreement among the five coders for each dimension.}
		\label{table_descriptives}
         \begin{tabular}{lrrr}
	\toprule
	\headrow Label        &             N & Fleiss' $\kappa$ &         Agreement \\ \midrule
	Anti-Elitism          &          3236 &            0.410 &           65.8 \% \\ \midrule
	People-Centrism       &          1608 &            0.244 &           81.8 \% \\ \midrule
	Left-Wing Ideology    &          1393 &            0.355 &           84.5 \% \\ \midrule
	Right-Wing Ideology   &           773 &            0.364 &           91.6 \% \\ \midrule
	\textbf{Total / Mean} & \textbf{8795} &   \textbf{0.343} & \textbf{80.93 \%} \\ \bottomrule
\end{tabular}
	\end{threeparttable}
\end{table}

Due to the absence of a strict external ground truth regarding populism, many labels allow different opinions that may be considered correct.
Accordingly, discussions with our annotators showed that, in particular, the condition of moralistic language generated disagreement (although all coders presented reasonable justifications for their judgments).
It became evident that the annotation process reflects reasonings on populism as a concept that eludes a strict definition \parencite{klamm-etal-2023-kind}.

This stands in contrast to most annotations in computational linguistics (CL), which are designed for rather objective tasks with the underlying assumption that each item \textit{can} be defined to be true (e.g., part-of-speech tagging). 
However, such a single, objective truth (sometimes called ``gold labels'') does not hold for more abstract and latent aspects of natural language \parencite{umaLearningDisagreementSurvey2021}.
Consequently, a growing branch of NLP researchers acknowledges that disagreement between coders should not be disregarded by aggregation strategies like majority voting \parencite{plankProblemHumanLabel2022}.\footnote{Focusing on limitations of human-coded annotations is a rather new development in the NLP community and is often labeled \textit{perspectivism}. See, for instance, \url{https://pdai.info/}.
One of the consequences is to release our data on the level of annotators to allow further investigations of annotators' (dis)agreement.}
We embrace this reasoning and argue that each labeled instance of our well-trained coders might reflect an aspect of populism.
Hence, if at least one coder identifies a dimension of populism in a sentence, we assume that there exists something worth learning for the model to capture as many facets of the concept as possible.

Additionally, we treat our classification problem as a multilabel problem, meaning each sentence may contain multiple dimensions simultaneously. 
Take the following sentence as an example: ``The consequences of your [the government] inaction are paid by society, paid by taxpayers, paid by farmers.''\footnote{Original sentence: Die Folgen Ihres Nichthandelns bezahlt die Gesellschaft, bezahlen die Steuerzahler, bezahlen die Bäuerinnen und Bauern. It was said by Friedrich Ostendorff (Grüne) on June 29, 2015.}
In this sentence, the elite is criticized with a moralizing undertone.
At the same time, the speaker aligns themselves with the people, represented by ``taxpayers'' and ``farmers,'' which provides compelling evidence to label it as people-centristic.
Thus, there would be no adequate way of choosing anti-elitism \textit{or} people-centrism. 

Using a multilabel model has another advantage: We augment our data on the populist dimensions by learning the host ideology separately.
In a multiclass model, we would need to create all combinations of our dimensions: left-wing anti-elite, neutral anti-elite, and right-wing anti-elite (same for people-centrism).
This way, the classification in the neutral anti-elite category would no longer be influenced by the samples that fall into either of the other two categories.
Using a multilabel approach, we resolve this division and learn the categories independently.

\section{Results}\label{sec:results}

To meet the requirements of the complex task of detecting populism in text, we present, on the one hand, common performance indicators for ML models, i.e., precision, recall, and summarizing F1-scores for each dimension.
On the other hand, we employ a battery of validity checks to (i) examine the intuition behind typical sentences for each dimension, (ii) aggregate sentences on the level of parties and speakers to compare our results to expert surveys and other classification models, and (iii) measure external validity by out-of-sample predictions.

\subsection{Model performance}

We used a train-validation-test split (60\%-20\%-20\%) to optimize the hyperparameters and to identify the optimal thresholds for each dimension.
Through an iterative process, we found that a batch size of 16 and a learning rate of 4e-6 combined with a cosine annealing learning rate scheduler that reduces the learning rate to 1e-9 over all epochs yielded good results.
Furthermore, an AdamW-optimizer with a weight decay of 1e-2 was used. 
We stopped training after 13 epochs as the validation loss reached a plateau. 
The thresholds for each dimension were selected to maximize the F1-score on the validation set.\footnote{Our threshold search found the following values to optimize F1-scores: .501 for anti-elitism, .502 for people-centrism, .422 for left-wing ideology, and .383 for right-wing ideology.}

\begin{table}
\begin{threeparttable}
\caption{Performance across a 5-fold cross-validation. The standard deviation across each run is reported in parentheses. Hyperparameters are consistent for each of the 5 splits.}
\label{table_classification_report}
\begin{tabular}{llll}
	\toprule
	\textbf{Dimension}  & \textbf{Precision} & \textbf{Recall} & \textbf{F1}   \\ \midrule
	Anti-Elitism        & 0.812 (0.013)      & 0.885 (0.006)   & 0.847 (0.007) \\ \midrule
	People-Centrism     & 0.670 (0.011)      & 0.725 (0.040)   & 0.696 (0.019) \\ \midrule
	Left-Wing Ideology  & 0.664 (0.023)      & 0.771 (0.024)   & 0.713 (0.010) \\ \midrule
	Right-Wing Ideology & 0.654 (0.029)      & 0.698 (0.050)   & 0.674 (0.031) \\ \midrule
	&                    &                 &               \\ \midrule
	micro avg           & 0.732 (0.009)      & 0.805 (0.006)   & 0.767 (0.007) \\ \midrule
	macro avg           & 0.700 (0.011)      & 0.770 (0.010)   & 0.733 (0.010) \\ \bottomrule
\end{tabular}
\end{threeparttable}
\end{table}

For evaluating the results, we employed 5-fold cross-validation. 
Hence, we used the aforementioned hyperparameter configuration to train five models, each of which relied on 80\% of the data as the training set and on the remaining 20\% as the test set. 
The results are shown in Table \ref{table_classification_report}.

Of the two main dimensions, the models' ability to detect anti-elitism is clearly higher than for people-centrism. 
This is in accordance with results reported by \textcite{klamm-etal-2023-kind}.
There may be two reasons: First, we found around twice as many instances of anti-elitism in the Bundestag corpus than people-centrism; secondly, the definition of people-centric relies more on individual perspectives (e.g., which groups qualify as instances of ``the people'').
However, even the lower F1-scores for people-centrism are in the same range as reported by \textcite{bonikowskiPoliticsUsualMeasuring2022}---albeit for different dimensions, but in a similar setup. 
While we do not have any comparable studies for detecting left-wing or right-wing populist ideology in text, the F1-scores of 0.713 and 0.674, respectively, represent no outliers from our main dimensions. 

A tendency across all dimensions in Table \ref{table_classification_report} is that recall is higher than precision. 
While false negatives (responsible for low recall) would be eliminated from a subsequent evaluation of the results and cannot be examined or adjusted, aggregating multiple predictions (for example, one prediction per sentence of a speech) or setting higher thresholds should reduce the number of false positives (responsible for low precision).
Since our classifier is not intended to decide based on a single prediction, we assume recall is more important for applications than precision.

A final takeaway of Table \ref{table_classification_report} is that the performance does not vary across the different splits.
We find similar performances across resampled groups, from which we infer that the performance does not rely on some particular advantageous split or setup.
Therefore, we trained the final model (PopBERT) on the complete data using the aforementioned configuration.

\subsection{Revealing populist language in the Bundestag}

Out of all 1,258,876 analyzed sentences from the 18th and 19th Bundestag, we find 6.7 percent to contain anti-elitism, 2.2 percent people-centrism, 1.1 percent a left-wing host ideology, and 0.6 percent a right-wing host ideology.\footnote{To calculate the percentages, we used each dimension's threshold (see above). If a sentence has a higher probability than the threshold, it is considered to contain one of the dimensions.}
We first selected sentences representing different dimensions and combinations to provide qualitative insights into how the model works (Table \ref{table_examples}).
While we have no ground truth in a strict sense, the predictions make intuitive sense (more examples based on a stratified random sample can be found in \ref{ap_example_sents}).

\begin{table}
	\begin{threeparttable}
		\caption{Selected examples. To demonstrate PopBERT's predictive behavior, three sentences have been selected that correspond to only one of the two core dimensions (\textit{anti-elitism} or \textit{people-centrism}) as well as three sentences containing both dimensions simultaneously, thus resembling the definition of \textit{populism} presented in Section \ref{sec:theory}. For each of these, one sentence is predicted to be neutral, one is attached to a left-wing, and one to a right-wing host ideology. However, it is important to note that these sentences have been selected manually. We present further examples of a stratified random sample in \ref{ap_example_sents}.}
		\label{table_examples}
		\begin{tabular}{llrrrrrr}
	\toprule
	\headrow Dimension                        & ID &                                                                                                                                                                                                                                                                                                                                                                                                                                          Text & \parbox[t]{0.6cm}{Gründl (2022)} &        elite &        centr &         left &        right \\ \midrule
	\parbox[t]{1cm}{\textbf{Anti-Elitism}}    & 1  &                                                                                                                                                                                                                                                                               \parbox[t]{6.5cm}{The opposition is already on an intellectual summer break.\\
		\textit{Die Opposition befindet sich intellektuell bereits in der Sommerpause.}} &                                0 & \textbf{.99} &          .00 &          .00 &          .07 \\ \midrule
	                                          & 2  &                                                                                                                                                                                             \parbox[t]{6.5cm}{The deregulation of financial markets is what initially sparked dollar signs in the eyes of speculators.\\
		\textit{Die Deregulierung der Finanzmärkte hat doch erst die Dollarzeichen in die Augen der Spekulanten gezaubert.}} &                                1 & \textbf{.97} &          .00 & \textbf{.93} &          .00 \\ \midrule
	                                          & 3  &                                                                                                                                                                         \parbox[t]{6.5cm}{In great haste, you stumble through the country with your policies, hastily passing one asylum package after another.\\
		\textit{In großer Hektik taumeln Sie mit Ihrer Politik durch das Land und beschließen Sie ein Asylpaket nach dem anderen.}} &                                0 & \textbf{.99} &          .01 &          .03 & \textbf{.96} \\ \midrule

	\parbox[t]{1cm}{\textbf{People-Centrism}} & 4  &                                                                                                                                                                                                                                                                 \parbox[t]{6.5cm}{We make concrete policy for the very concrete problems of the people.\\
		\textit{Wir machen konkrete Politik für die ganz konkreten Probleme der Menschen.}} &                                0 &          .00 & \textbf{.99} &          .02 &          .01 \\ \midrule
	                                          & 5  &                                                                                                                               \parbox[t]{6.5cm}{We do not demand any social policy measures for these people, but we demand: Stop the permanent social exclusion!\\
		\textit{Wir fordern für diese Menschen nicht irgendwelche sozialpolitischen Maßnahmen, sondern wir fordern: Schluss mit der dauerhaften gesellschaftlichen Ausgrenzung!}} &                                0 &          .00 & \textbf{.98} & \textbf{.96} &          .00 \\ \midrule
	                                          & 6  &                                                                              \parbox[t]{6.5cm}{A Syrian with 4 wives and 23 children already costs the German taxpayer 400,000 euros per year in support, without ever having contributed a single cent.\\
		\textit{Ein Syrer mit 4 Frauen und 23 Kindern kostet den deutschen Steuerzahler schon jetzt im Jahr 400.000 Euro Alimentierung, ohne je einen einzigen Cent eingezahlt zu haben.}} &                                0 &          .24 & \textbf{.95} &          .08 & \textbf{.99} \\ \midrule

	\textbf{Populism}                         & 7  &                                                                                                                                                                                                                                                           \parbox[t]{6.5cm}{For tenants, this government inaction is really costing them dearly.\\\textit{Für Mieterinnen und Mieter kommt diese Tatenlosigkeit der Regierung wirklich teuer.}} &                                0 & \textbf{.99} & \textbf{.76} &          .01 &          .01 \\ \midrule
	                                          & 8  &                                                                         \parbox[t]{6.5cm}{This is class warfare from above; this is class warfare in the interests of the wealthy and the propertied against the majority of taxpayers on this earth.\\
		\textit{Das ist Klassenkampf von oben, das ist Klassenkampf im Interesse von Vermögenden und Besitzenden gegen die Mehrheit der Steuerzahlerinnen und Steuerzahler auf dieser Erde.}} &                                1 & \textbf{.85} & \textbf{.99} & \textbf{.98} &          .20 \\ \midrule
	                                          & 9  & \parbox[t]{6.5cm}{You are the ones who are widening the gender pay gap by trampling upon poor German female retirees, while lavishly giving money to asylum seekers, who are predominantly young and male.\\
		\textit{Sie sind jene, die den Gender Pay Gap vergrößern, indem Sie die armen deutschen Rentnerinnen mit Füßen treten, während Sie den Asylbewerbern, die hauptsächlich jung und männlich sind, das Geld in den Rachen werfen.}} &                                0 & \textbf{.98} & \textbf{.98} &          .11 & \textbf{.99} \\ \midrule
\end{tabular}
	\end{threeparttable}
\end{table}

Consider the examples of anti-elitism (sentences 1--3). 
As described by theory, each of them addresses parts of the perceived elite in a moralizing way (i.e., opposition parties, speculators, and ``your policies'').
Sentence 2 is also attached to the left-wing ideology since the criticized group is part of the financial elite, which is one of the main antagonists of the political left. 
In contrast, sentence 3 criticizes the government's execution of the asylum policy, a typical link made by right-wing ideologists.
Accordingly, the sentences illustrating people-centrism address either ``the(se) people'' directly (sentences 4 and 5), or representative parts like ``the German taxpayer'' in sentence 6.
Also following theoretical guidelines, the sentence classified as left-wing addresses social deprivation, while the right-wing host ideology policy points to immigrants.
The last group of examples shows instances that contain both dimensions, i.e., addresses a perceived elite in a pejorative way and makes a claim for the people (or some explicit part of those, e.g., ``tenants'', ``taxpayers on earth'', or ``German female retirees'').
Again, the model marks sentences revolving around social class as left-wing and the statement that seeks to blame immigrants as right-wing.

\begin{table}
\begin{threeparttable}
    \caption{Most populist politicians in the German Bundestag for the 18th (2013--2017) and 19th (2017--2021) electoral term. The ranking rests on a multiplicative index of speeches' probabilities for anti-elitism and people-centrism. Both dimensions must be present to meet the definition of populism (cf. Section \ref{sec:theory}). We first calculated the dimension-specific mean across each politician's speech in the corpus for that purpose. We then multiplied the results of both dimensions.}\label{table_politicians}
	\begin{tabular}{lllll}
	\toprule
	\headrow Term & Rank & First Name & Last Name   & Party     \\ \midrule
	\textbf{18}   & 1    & Sahra      & Wagenknecht & DIE LINKE \\ \midrule
	              & 2    & Katja      & Kipping     & DIE LINKE \\ \midrule
	              & 3    & Sabine     & Zimmermann  & DIE LINKE \\ \midrule
	              & 4    & Andreas    & Scheuer     & CDU/CSU   \\ \midrule
	              & 5    & Jutta      & Krellmann   & DIE LINKE \\ \midrule
	              &      &            &             &           \\ \midrule
	\textbf{19}   & 1    & Martin     & Sichert     & AfD       \\ \midrule
	              & 2    & Alice      & Weidel      & AfD       \\ \midrule
	              & 3    & Martin     & Reichardt   & AfD       \\ \midrule
	              & 4    & Bernd      & Riexinger   & DIE LINKE \\ \midrule
	              & 5    & Gottfried  & Curio       & AfD       \\ \bottomrule
\end{tabular}
\end{threeparttable}
\end{table}

However, many important questions address more aggregated levels of texts, for instance, analyzing the language used by persons or parties. 
To trace PopBERT's construct validity in this respect, we first aggregate at the level of politicians and create a ranking of the most populist politicians in the Bundestag.
As described in Section \ref{sec:theory}, many scholars \parencite[e.g.,][]{ernstBipolarPopulismUse2017, daiWhenPoliticiansUse2022} argue that it is the combination of both dimensions, anti-elitism and people-centrism, that constitutes populist statements.
To meet this definition, we propose a multiplicative index.
We therefore take the average over all dimensions and phrases of a politician during a legislative period and multiply both averages per politician.
This ensures that higher values in both dimensions also result in a higher index value, while the absence of at least one dimension yields an index value of 0.
Furthermore, it penalizes textual units where the dimensions are distributed very unevenly compared to units with a more equal distribution over both dimensions.

Table \ref{table_politicians} presents politicians with the highest propensity for using populist language in the Bundestag.
While an in-depth analysis is beyond the scope of this paper, the table indicates two notable trends: 
First, the entry of the AfD into the Bundestag brought about a significant shift.
In the 18th term, members from DIE LINKE dominated the list.
Since entering in 2017, however, the AfD members clearly took the lead in using populist language. 
Secondly, the top-ranking individuals in both terms are prominent public figures known for their provocative statements.
These figures include, for instance, Sarah Wagenknecht (DIE LINKE), who has authored several controversial books and has a substantial followership on social media platforms (over 680,000 followers on Twitter and 660,000 on YouTube, respectively).
In the 19th term, Alice Weidel was among the most populist politicians.
She is also a controversial political figure with a significant social media presence(270,000 followers on Twitter and 350,000 likes on Facebook).

\begin{figure*}[!htb]
	\centering
	\includegraphics[width=\linewidth]{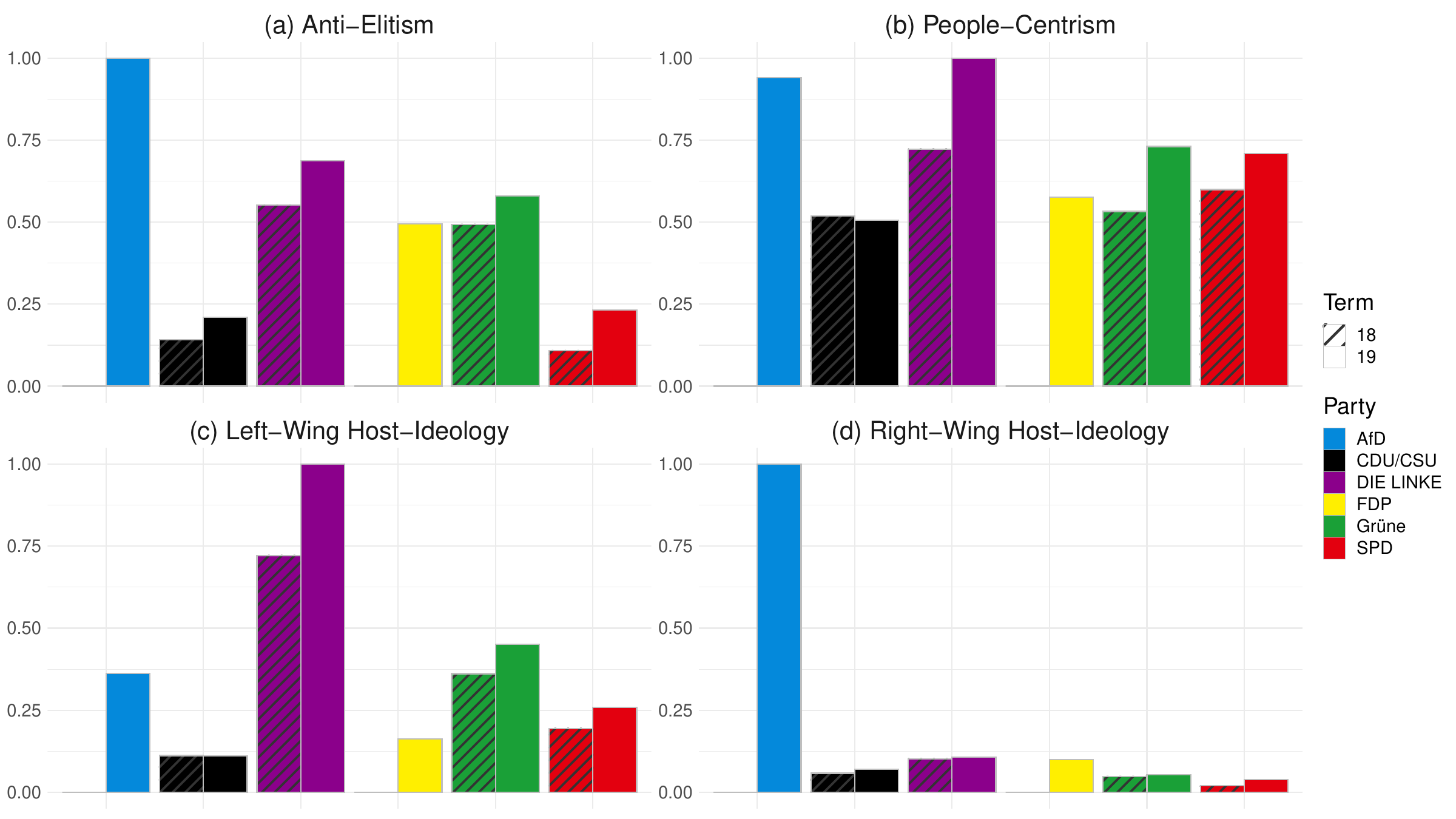}
	\caption{Populist dimensions in speeches of the 18th and 19th legislative period of the Bundestag, by party. This figure illustrates the model predictions for all four dimensions averaged across all sentences in the dataset, per party. The values are normalized to the respective maximum value of each dimension to highlight the proportions between the parties. Subplots with unstandardized values can be found in \ref{ap_orig_figures}.}
	\label{fig_all_dimensions}
\end{figure*}

To further explore whether PopBERT classifies populist language according to human judgment, we aggregate sentences by parties.
Figure \ref{fig_all_dimensions} presents the mean values of populism for each dimension by party.\footnote{To avoid strong influence from very short speech contributions, we exclude all contributions that contain fewer than four sentences.} 
Each dimension confirms the expected results: The AfD demonstrates by far the highest values in the right-wing and anti-elitism dimensions, while politicians from DIE LINKE use much more left-wing populist sentences than the other parties.
Interestingly, DIE LINKE exhibits higher values for people-centrism than other political parties.
In general, the distribution of people-centrism across all parties is more balanced than anti-elitism.
Furthermore, we observe increased values in the 19th legislative period.
However, a more detailed analysis of a potential rise in populism is beyond the scope of this paper.
It is, for instance, possible that the later period differs systematically in linguistic aspects from the previous one.

Next, we compare PopBERT's predictions with the Chapel Hill Expert Survey \parencite[CHES;][]{jollyChapelHillExpert2022}.
This survey measures two dimensions: ``antielite\_salience,'' which aligns with our concept of anti-elitism, and ``people\_vs\_elite,'' which \textit{roughly} corresponds to the definition of people-centrism outlined above.
Due to the differing temporal coverage of the CHES data, we juxtapose the data collected in the year 2019 with the values we have calculated for the 19th term.
We find a strong Pearson correlation coefficient for anti-elitism ($r=.896$) and people-centrism ($r=.746$), indicating that our results align well with the CHES expert ratings.

\begin{figure*}[!htb]
	\centering
	\includegraphics[width=\linewidth]{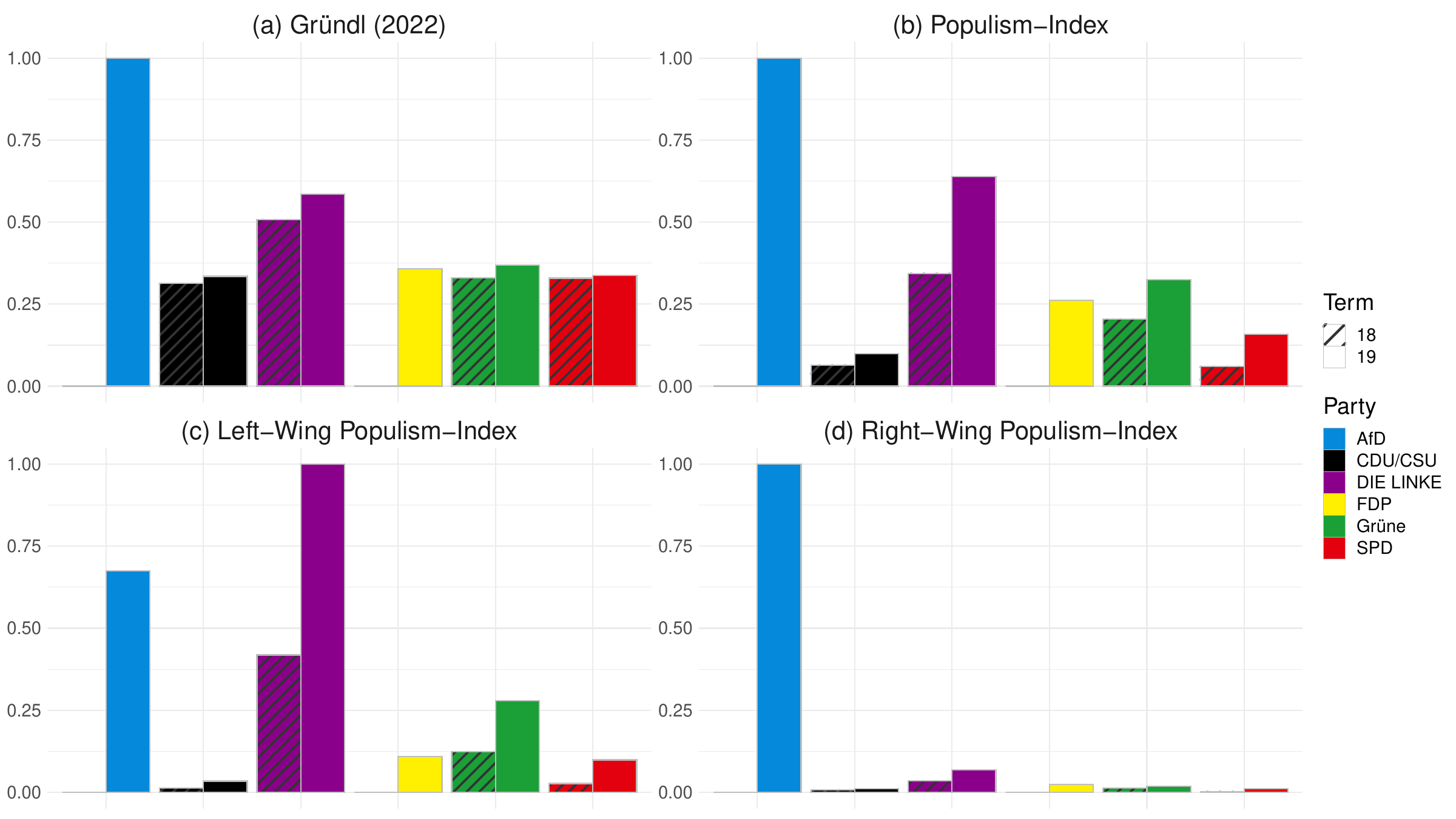}
	\caption{Populism in speeches of the Bundestag's 18th and 19th legislative period, by party. Subplot (a) illustrates populism according to the method proposed by \textcite{grundl_populist_2022}. (b) shows the predicted values of the multiplicative populism index. (c) and (d) display the results for left-wing and right-wing populism, respectively. All values in each subplot have been scaled to their respective maximum values so that the bars can be interpreted as proportional to the maximum value of each dimension.}
	\label{fig_multi_index}
\end{figure*}

Figure \ref{fig_multi_index} displays the multiplicate populism index (i.e., considering the product of anti-elitism and people-centrism) per party and speech in the 18th and 19th Bundestag.
We observe the highest manifestation of populism in speeches of the AfD, followed by DIE LINKE, while the governing parties are found at the lower end of the spectrum.
Similar to the results in Figure \ref{fig_all_dimensions} on ideology (i.e., only being attached to either anti-elitism or people-centrism), we find that the AfD is almost the only party in the 19th Bundestag using right-wing populist language. 
While DIE LINKE is also most associated with left-wing populist speeches, we still see considerable left-wing populism from the AfD and Greens. 

Additionally, we compare our method to the approach of  \textcite{grundl_populist_2022}.
Although he did not intend to identify populist language in parliamentary speeches but in social media posts, his dictionary is the only other classifier of German populist speech.
The upper left panel of Figure \ref{fig_all_dimensions} shows that the dictionary also ranks the AfD and DIE LINKE as the most populist parties. 
However, the dictionary detects no differences between the other parties. 
The lack of nuances becomes even more apparent when we return to Table \ref{table_examples}.
Most examples there would not have been considered populist by a dictionary approach.

One of the main challenges for developing a language model to detect populism in texts---as it is for many complex phenomena \parencite{plankProblemHumanLabel2022}---is the lack of a binary answer as to whether a label or prediction is correct.
As a final step, we therefore searched the literature for ``true'' populist statements.
The reasoning is that if experts on populism claim that a statement is a typical case, this represents the best approximation of a ``true'' populist utterance.
We identified four studies that provide 17 prototypical examples of populist statements from Facebook posts \parencite{schurmannYellingSidelinesHow2022, ernstExtremePartiesPopulism2017} and presidential campaign speeches \parencite{bonikowskiPoliticsUsualMeasuring2022,daiWhenPoliticiansUse2022}.
We use these text snippets for an out-of-sample prediction.
In \ref{apextval}, we report the respective wording, intended dimension, and prediction probabilities for PopBERT's four dimensions.
For 16 out of 17 sample sentences (94.1\%), the classifier predicts at least one of the core dimensions. 
The single exception is a rather idiosyncratic sentence.
Even though each example's wording, style, and length differs considerably, the results strongly support the notion that PopBERT is able to detect different dimensions of populism in out-of-sample texts.

\section{Discussion}

This paper introduces PopBERT, a transformer-based language model to identify populist utterances.
The model was trained on speeches from the German parliament spanning 2013 to 2021. 
Using a widely accepted theoretical framework, we annotated two key dimensions of populism: anti-elitism and people-centrism.
Additionally, we labeled instances in which left-wing or right-wing host ideologies are attached to the ``thin ideology'' of populism \parencite{muddePopulistZeitgeist2004}.
The resulting multilabel classifier demonstrated high precision and, in particular, recall.
We further assessed the model's concept validity through out-of-sample performance and its correspondence to intuitive and expert judgments at different aggregation levels (sentence, speaker, party).

Various potential applications of PopBERT can be envisioned.
For instance, it may allow researchers to identify rhetorical patterns, contextual factors, or temporal dynamics underlying populist language.
Moreover, PopBERT enables examining how populists frame their messages and which social, economic, or cultural issues resonate in their language.
In so doing, information on the personal characteristics of the politicians might also contribute to current questions in the CL community \parencite{hovyImportanceModelingSocial2021}.
Those calls ask for a closer inspection of the interplay between social-demographic attributes and language usage.
Studies merging personal characteristics with political content---such as the correlation between politicians' gender and their prevalent issues \parencite{backWhenWomenSpeak2019}---already exemplify promising avenues for future research.
Ultimately, PopBERT's ability to identify populist statements, investigate their content, and link them to personal attributes may shed light on the relationship between the success of populists and their language usage.

Publishing a detailed codebook and annotator-level labels might also contribute to methodological advances in detecting complex and ambiguous political concepts in language.
For instance, researchers could apply the model to a different dataset in a cross-domain classification setting and use it, e.g., to identify populist social media posts or populist language in newspapers.
It would also be straightforward to enrich the dataset with only a small amount of additional domain-specific data and train a customized model according to specific needs.

In the wake of the emerging \textit{perspectivism} methodology in CL \parencite{plankProblemHumanLabel2022}, we would also like to encourage researchers to analyze systematic variations in the coding behavior exhibited by our annotators.
Our five well-trained coders' (dis-)agreement could shed further light on which models work best for social constructs such as populism.
More elaborate model architectures or a more complex definition of the target labels, such as soft labels 
\parencite{umaCaseSoftLoss2020}, seem to be promising future research opportunities.
Finally, annotations and models might be easily updated with additional labels for existing dimensions or combined with complementary classifiers like, for example, developed by \textcite{klamm-etal-2023-kind}.

Those potential applications and extensions notwithstanding, measuring populist language faces important challenges.
Its ambiguity yields agreement rates similar to other studies modeling rather complex phenomena like emotions or morality \parencite{kobbe-etal-2020-exploring, wood-etal-2018-comparison}.
However, it contrasts the work of \textcite{bonikowskiPoliticsUsualMeasuring2022}, who relied on two coders and utilized a third decision-maker in ambiguous cases.
This procedure yields, by design, higher agreement rates.
The trade-off lies in an increased variety of cases versus higher agreement.
We decided to favor the first option because we argue that each labeled instance of our well-trained coders helps the model to find nuances of populism, and utilizing a multilabel model augments the available training data.
It is an open question, though, which way should be preferred and for what cases. 
Future comparisons of different methodologies across the \textit{same} concept should help develop important---and, to a large extent, still missing---guidelines for large-scale language models.

While we intentionally chose sentence-level annotations to create a flexible model with broad applicability, we are aware that longer text segments would have likely reduced disagreement among coders.
However, labeling segments beyond the clear-cut grammatical unit of sentences opens a box of other problems, e.g., the comparability of text snippets or the generalizability of model predictions.
Selecting text ranges is always a compromise between model applicability and interpretability.
We decided to favor applicability using the sentence level because, in so doing, other researchers can aggregate the predictions to contexts of their choosing such as paragraphs, speeches, politicians, or parties.
Furthermore, our sampling strategy using active learning has yielded a time-specific sample.
Therefore, we cannot evaluate how the model performs in other periods.
These limitations notwithstanding, PopBERT and its accompanying dataset may provide valuable tools for studying populism in German texts, opening further ways of interdisciplinary research, and contributing to a deeper understanding of the causes and effects of populist language. 

\begin{acknowledgement}
	We want to thank Sophia Hunger, Lukas Wertz, Marius Kaffai, and the participants of the SICSS Munich for their insightful comments.
\end{acknowledgement}

\paragraph{Data Availability Statement} Replication data to replicate all figures and tables in this article is available at Erhard (2023) \url{https://doi.org/10.7910/DVN/HZMSUR}. This repository also holds a version of the manually labeled data. The model installation and usage instructions are available at \url{https://huggingface.co/luerhard/PopBERT}. Python and R-Code used during the analysis and creation of the model are available at \url{https://github.com/luerhard/PopBERT}.

\paragraph{Funding Statement} This research was supported by grants from the Deutsche Forschungsgemeinschaft (DFG, German Research Foundation under the DFG reference number \textbf{UP 31/1}).

\paragraph{Competing Interests} None.


\printbibliography

\newpage
\appendix

\begin{minipage}{\textwidth}
	\centering
	\textbf{\large Online Appendix to: PopBERT. Detecting populism and its host ideologies in the German Bundestag}
	\vspace{2cm}
	
	Lukas Erhard, University of Stuttgart\\
	Sara Hanke, University of Stuttgart\\
	Uwe Remer, University of Stuttgart\\
	Agnieska Falenska, University of Stuttgart\\
	Raphael Heiberger, University of Stuttgart
	
\end{minipage}

\newpage
\section{Dataset description}\label{ap_dataset}

\subsection{Full Dataset}\label{ap_full_dataset}

\begin{table}[H]
	\centering
	\begin{tabular}{llrrll}
\toprule
Term & Party & Speeches & Sentences & Avg. sentences & Std. sentences \\\midrule
\multirow[t]{5}{*}{18} & CDU/CSU & 4,437 & 251,109 & 56.594 & 23.651 \\
 & SPD & 3,397 & 177,056 & 52.121 & 21.806 \\
 & Grüne & 2,831 & 116,538 & 41.165 & 27.980 \\
 & DIE LINKE & 2,444 & 108,251 & 44.293 & 29.499 \\
 & Fraktionslos & 2 & 52 & 26.000 & 2.828 \\
\multirow[t]{7}{*}{19} & CDU/CSU & 5,104 & 225,145 & 44.111 & 16.806 \\
 & SPD & 3,604 & 150,928 & 41.878 & 17.134 \\
 & AfD & 3,020 & 109,504 & 36.260 & 17.293 \\
 & FDP & 2,622 & 93,045 & 35.486 & 18.412 \\
 & Grüne & 2,495 & 88,244 & 35.368 & 18.337 \\
 & DIE LINKE & 2,321 & 84,132 & 36.248 & 20.005 \\
 & Fraktionslos & 71 & 2,022 & 28.479 & 7.107 \\\midrule
sum &  & 32,348 & 1,406,026 &  &  \\
\bottomrule
\end{tabular}
	\caption{Number of sentences and speeches in the data set by electoral term and party. \textit{Avg. sentences} depict the mean number of sentences per speech, party, and term. \textit{Std. Sentences} shows the corresponding standard deviations.}
\end{table}

\subsection{Distribution of annotated dimensions}\label{ap_label_dataset}

\begin{table}[H]
	\begin{threeparttable}
		\begin{tabular}{llrrrrrr}
\toprule
 &  & antielite & pplcentr & eliteless & pplmore & left & right \\
Term & Party &  &  &  &  &  &  \\
\midrule
\multirow[t]{5}{*}{18} & CDU/CSU & 339 & 245 & 13 & 7 & 92 & 78 \\
 & DIE LINKE & 976 & 256 & 13 & 59 & 617 & 34 \\
 & Fraktionslos & 12 & 0 & 0 & 0 & 0 & 0 \\
 & Grüne & 588 & 178 & 7 & 35 & 275 & 16 \\
 & SPD & 199 & 244 & 1 & 19 & 151 & 14 \\
\midrule
\multirow[t]{7}{*}{19} & AfD & 2,510 & 521 & 38 & 69 & 294 & 1,097 \\
 & CDU/CSU & 452 & 243 & 4 & 25 & 69 & 84 \\
 & DIE LINKE & 1,044 & 348 & 6 & 33 & 721 & 15 \\
 & FDP & 408 & 154 & 6 & 10 & 44 & 48 \\
 & Fraktionslos & 101 & 18 & 8 & 10 & 3 & 35 \\
 & Grüne & 676 & 220 & 3 & 29 & 279 & 14 \\
 & SPD & 405 & 288 & 1 & 23 & 205 & 17 \\
\midrule
sum &  & 7,710 & 2,715 & 100 & 319 & 2,750 & 1,452 \\
\bottomrule
\end{tabular}
		\caption{Number of labels in the data set by electoral term, party, and whether a coder was unsure. These numbers represent sums over all coders, so each labeled sample is present multiple times (once per coder).} 
	\end{threeparttable}
\end{table}

\section{Codebook} \label{sec:codebook}
The following paragraphs describe the final codebook that was given to the annotators after several rounds of improvements to the described text. 
It is in German as all annotators and annotations were carried out in German as well. 
It contains descriptions of the used concept of (left-wing and right-wing) populism as well as examples following the definition outlined in our conceptualization of populism \ref{sec:theory} and is inspired in wide parts by the operationalization by \textcite{wirthAppealPopulistIdeas2016, ernstBipolarPopulismUse2017}.

\subsection*{Definition von Populism}
Wir definieren Populismus nach \textcite{muddePopulistZeitgeist2004}. Populismus ist ein Konzept, das drei zentrale Komponenten hat: \textit{Anti-Elitismus}, \textit{Volkszentrismus} und ein Fokus auf Volkssouveränität. Es wird eine widerstreitende Beziehung zwischen den korrumpierten Eliten (Anti-Elitismus) und dem tugendhaften Volk (Volkszentrismus) angenommen. Für Populisten soll Politik ein Ausdruck des Volkswillens sein, der als homogen verstanden wird.

Populismus lässt sich als „thin-centered ideology“ verstehen. Das heißt es handelt sich um eine Reihe von Ideen, die aber allein keine eigene Ideologie darstellen, aus der sich eine inhaltliche Richtung im politischen Spektrum ableiten lässt. Stattdessen können sie sich mit einer sogenannten „host (oder thick) ideology“ wie zum Beispiel Nativismus (Rechtspopulismus), Kommunismus/Sozialismus (Linkspopulismus), etc. verbinden. 

Es reicht nicht aus, dass in einem Text das Volk oder die Eliten genannt werden. Die Nennungen von Volk und die Eliten müssen mit wertenden Aussagen verbunden sein, z.B. normative Aussagen (so sollte es sein) oder moralische Aussagen (das ist richtig oder falsch). Nur dann kann die Aussage als populistisch angesehen und kodiert werden.

Das bedeutet, die Elite wird negativ beschrieben. Die Eliten werden beschuldigt, bösartig, kriminell, faul, dumm, extremistisch, rassistisch, undemokratisch usw. zu sein. Die Elite wird beschimpft und ihr werden Moral, Charisma, Glaubwürdigkeit, Intelligenz, Kompetenz, Beständigkeit usw. abgesprochen. Das Volk wird als positiv beschrieben. Das Volk ist mit Moral, Charisma, Glaubwürdigkeit, Intelligenz, Kompetenz, Beständigkeit usw. ausgestattet. Das Volk ist nicht böswillig, kriminell, faul, dumm, extremistisch, rassistisch, undemokratisch.

Keine populistische Aussage ist zum Beispiel:
\textit{„Die Regierung hat einen Fehler begangen.“}

Eine populistische Aussage ist zum Beispiel:
\textit{„Die Merkel-Diktatur hat einen Fehler begangen.“}

Diese beiden unterscheiden sich vor allem in einem moralisierenden Unterton, der in dem Wort „Merkel-Diktatur“ mitschwingt.

Kodieren Sie nur solche Aussagen als populistisch, die so gemeint sind. Wird deutlich, dass die Sprecher*in jemandem Populismus vorwirft oder jemanden zitiert wird dies nicht kodiert.
Das wird nicht codiert, weil es sich nicht um die Meinung der Sprecher*in, sondern die einer dritten Person handelt:
\textit{„Bernd Höcke behauptet, die Merkel-Diktatur mache viele Fehler.“}

Folgendes Beispiel wird allerdings trotzdem codiert, weil keine Schlussfolgerung über die Meinung des Sprechers möglich ist: 
\textit{„Der Film dokumentiert die Gier der Spekulanten und das Versagen der Politik.“}

\subsection*{Anti-Elitismus}
Unter Anti-Elitismus fallen alle Aussagen, wo die \textit{Eigenschaften} der Elite kritisiert werden (1), sie für eine \textit{schlechte Entwicklung/Situation} verantwortlich gemacht werden (2), und kritisiert wird, dass die Eliten \textit{nicht das Volk repräsentieren} (3).

(1)	Die Eigenschaften der Elite werden als negativ beschrieben. Die Eliten werden beschuldigt, bösartig, kriminell, faul, dumm, extremistisch, rassistisch, undemokratisch usw. zu sein. Die Elite wird beschimpft und ihr werden Moral, Charisma, Glaubwürdigkeit, Intelligenz, Kompetenz, Beständigkeit usw. abgesprochen.

(2)	Der Elite wird Schuld für eine Situation zugesprochen; entweder weil sie für die Herbeiführung der schlechten Situation verantwortlich gemacht oder weil sie für die Lösung der schlechten Situation keine Antworten hat. Eliten werden als Bedrohung/Belastung beschrieben, als verantwortlich für negative Entwicklungen/Situationen oder als diejenigen, die Fehler oder Verbrechen begangen haben. Eliten werden nicht als Quelle der Bereicherung oder als verantwortlich für positive Entwicklungen/Situationen beschrieben.

(3)	Die Beschreibung der Elite beinhaltet häufig, dass sie nicht zum Volk gehören, dem Volk nicht nahestehen, das Volk nicht kennen, nicht für das Volk sprechen, sich nicht um das Volk kümmern oder alltägliche Aufgaben nicht ausführen.

\subsubsection*{Beispiele}
\begin{itemize}
	\item  „Die deutsche Bundesregierung trägt eine Mitschuld an dieser Gräueltat.“
	\item  „Noch nie hat ein Ausschussvorsitzender so selbstherrlich und willkürlich versucht, die Opposition in einem Untersuchungsausschuss zum Schweigen zu bringen.“
	\item „Liebe SPD und Grüne, Sie sind hier besonders scheinheilige Vertreter; denn Sie müssten eigentlich kleinlaut zugeben, dass Sie im Vorfeld in Thüringen einen bürgerlichen Ministerpräsidenten verhindert haben.“
	\item „Sie sagen, Sie wollen, dass sich jeder in diesem Land an Recht und Ordnung hält - das wollen wir doch alle - , aber selber beschäftigen Sie eine ganze Reihe verurteilter Straftäter vom Geld der Steuerzahlerinnen und Steuerzahler in Ihren Reihen, in Ihrer Fraktion, in Ihren Büros.“
	\item „Sie geben also zu, dass Sie die deutschen Bürger und Steuerzahler auf eine über 500 Milliarden Euro teure Irrfahrt geführt haben, deren Zukunft eine  - ich wiederhole mich hier  - offene Baustelle ist.“
	\item „Wenn man sich die Bilder von der griechisch-türkischen Grenze anschaut, wo Menschenrechte mit Füßen getreten werden, wo mit Tränengas und Blendgranaten und auch scharf auf Flüchtlinge geschossen wird, dann schämt man sich, dass die Europäische Union so schändlich mit Menschen in Not umgeht.“
	\item „Das ist nichts für Hinterzimmer, nichts für Spielchen, nichts für unappetitliche Kuhhandel.“
	\item „In der Realität der Altparteien gab es dagegen jahrzehntelang überhaupt keine Regelungen, die wenigstens verhinderten, dass Regierungsmitglieder und Parlamentarische Staatssekretäre ihr amtlich erworbenes Wissen, vor allem aber ihre amtlich geknüpften Kontakte mehr oder weniger nahtlos nach dem Ausscheiden aus dem Amt gewinnbringend an den Meistbietenden verhökerten.“
\end{itemize}

Anti-Elitismus kann auch zusammen mit einer „host ideology“ vorkommen, wenn die Elite als nicht-nativ oder ökonomisch beschrieben wird. Mehr Informationen dazu finden Sie unter Punkt 1.4.1. Dann vergeben Sie den Code Anti-Elitismus + ideologische Ausrichtung. Hier sind einige Beispiele dazu:

\paragraph{Links}

\begin{itemize}
	\item „Es geht darin insbesondere um den Finanzmarkt, weil ein ganz relevanter Teil des Finanzmarktes leider auf Lug und Betrug basiert .“
	\item „Ihre Politik dürfte vielen Großkonzernen und Superreichen ganz recht sein.“
	\item „Es ist der zehnte Jahrestag der Finanzkrise, und die Krisenkosten haben eben nicht die 1 Prozent Superreichen und auch nicht die kriminellen Banker bezahlt, sondern die Menschen, die ihr Geld auf ehrliche Weise verdienen.“
	\item „Das ist ein politischer Kuhhandel im Interesse der Bauwirtschaft, ohne nennenswerten Nutzen für die Mieterinnen und Mieter und zulasten der Kommunen.“
	
\end{itemize}

\paragraph{Rechts}
\begin{itemize}
	\item „Unsere überschuldeten Haushalte stützen mit Milliardentransfers korrupte Regierungen und erzeugen neue Migration.“
\end{itemize}

\paragraph{Spezielles Beispiel}
\begin{itemize}
	\item „Die Bundesregierung feiert eine hyperkeynesianische Party – Sie haben es schon angedeutet, ich füge ‚hyper‘ hinzu– und verkonsumiert die Früchte des künstlichen Booms über fehlgeleitete Programme, mehr Zuwanderung, mehr Euro-Rettung, mehr Türkei-Hilfen, mehr Supranationalismus und damit gegen Deutschland.“
\end{itemize}

Wie Sie gemerkt haben, lässt sich hier eindeutig linke und rechte host ideology finden, wenn so etwas vorkommt, dann und nur dann codieren Sie beide host ideologies (links \& rechts). Wenn Sie nicht sicher sind, dann codieren Sie keine „host ideology“.

\subsection*{Volkszentrismus}
Unter Volkszentrismus fallen alle Aussagen, wo die \textit{Eigenschaften} des Volks gelobt werden (1), es für eine \textit{positive Entwicklung/Situation} verantwortlich gemacht wird (2), das Volk als \textit{einheitlich und homogen} beschrieben wird (3) und der Sprecher sich als dem Volk zugehörig beschreibt (4).

(1)	Die Eigenschaften des Volks werden als positiv beschrieben. Das Volk ist mit Moral, Charisma, Glaubwürdigkeit, Intelligenz, Kompetenz, Beständigkeit usw. ausgestattet. Das Volk ist nicht böswillig, kriminell, faul, dumm, extremistisch, rassistisch, undemokratisch.

(2)	Dem Volk wird Verantwortung für eine positive Situation zugesprochen. Das Volk wird als eine Bereicherung oder als verantwortlich für eine positive Entwicklung/Situation beschrieben. Das Volk wird als keine Bedrohung/Belastung, als nicht verantwortlich für negative Entwicklungen/Situationen beschrieben und nicht als hätte es Fehler oder Verbrechen begangen. 

(3)	Die Beschreibung des Volkes zeichnet sich durch gemeinsame, homogene Gefühle, Wünsche oder Meinungen aus.

(4)	Der Sprecher beschreibt sich selbst als zum Volk gehörend, dem Volk nahestehend, das Volk kennend, für das Volk sprechend, sich um das Volk kümmernd, mit dem Volk übereinstimmend oder alltägliche Aufgaben ausführend. Der Redner behauptet, das Volk zu vertreten oder zu verkörpern.

\subsubsection*{Beispiele}
\begin{itemize}
	\item „Das heißt, Sie haben noch eine halbe Stunde Zeit, sich ein Rückgrat wachsen zu lassen und sich für die gemeinsame Sache unserer Bürger und den Rechtsstaat einzusetzen.“
	\item „Es ist jedenfalls nicht gemeinwirtschaftlich, wie Sie schreiben, die Kosten einer Minderheit auf die Mehrheit der Bevölkerung abzuwälzen.“
\end{itemize}

Volkszentrismus kann auch zusammen mit einer „host ideology“ vorkommen, wenn das Volk als nativ oder marginalisiert beschrieben wird. Mehr Informationen dazu finden Sie unter Punkt 1.5.2. Dann vergeben Sie den Code Volkszentrismus + ideologische Ausrichtung. Hier sind einige Beispiele dazu:

\paragraph{Links}
\begin{itemize}
	\item „Wenn Einwanderungswillige lediglich als Arbeitskräfte betrachtet werden und nicht als Menschen, wenn die Familien nicht mitgedacht werden, wenn keine langfristigen Perspektiven geboten werden und wenn die Migrantinnen und Migranten als Arbeitskräfte zweiter Klasse behandelt werden, wird es Deutschland nie vom Einwanderungsland zu einer inklusiven und chancengerechten Einwanderungsgesellschaft schaffen.“
	\item „Sie sanieren den Haushalt auf dem Rücken der Ärmsten, und das ist aus unserer Sicht nicht hinnehmbar.“
\end{itemize}

\paragraph{Rechts}
\begin{itemize}
	\item „Schluss mit der Spaltung des deutschen Volkes und der Herrschaft des Unrechts – für Einigkeit und Recht und Freiheit für das deutsche Vaterland.“
\end{itemize}

\paragraph{Kein Volkszentrismus wäre z.B.}
\begin{itemize}
	\item „Durch Zuwanderung wird dieses Problem massiv verschärft.“
	\item „Migration ist die Ursache aller Probleme.“
	\item  „Unser Ziel ist, mehr Menschen mit Behinderungen in Ausbildung und Arbeit zu bringen.“
\end{itemize}
Weil hier kein Bezug auf „das Volk“, „die Bürger“ etc genommen wird, im letzten Satz findet sich keine moralische Aussage und ist deshalb nicht als populistisch zu werten.

\subsection{Links-Rechts}
Wie beschrieben, kann sich Populismus mit einer „host ideology“ verbinden. Diese kann sich auf Klasse (Links) oder Nativismus (Rechts) beziehen. Populismus kann, muss aber nicht mit einer anderen Ideologie verbunden auftreten.
Deshalb werden diese Typen in getrennt voneinander kodiert. Es wird daher 
\begin{enumerate}
	\item zunächst die Populismus-Dimension kodiert
	\item  wenn sie dann bestimmen können, ob es sich bei der Aussage um eine linke oder rechte Aussage handelt, vergeben sie zusätzlich noch den entsprechenden Code. Wenn sie nicht bestimmen können, ob es eine linke oder rechte Aussage ist, vergeben sie keinen Code.
\end{enumerate}

\subsubsection*{Links-Rechts - Verständnis von Elite}
\paragraph{Links} 
Die ökonomischen Eliten werden als Stütze der politischen Eliten gesehen; oder die politischen Eliten als von wirtschaftlichen Eliten unterwandert/korrumpiert. 
Sowohl wirtschaftliche Eliten per se (internationale Wirtschaftsorganisationen z.B. WTO, IWF, Weltbank; Unternehmen, etc) als auch die von wirtschaftlichen Interessen, dem Kapitalismus oder Neoliberalismus geleiteten Politiker werden kritisiert. 

\paragraph{Rechts}
Nicht-Native werden als Stütze der politischen Eliten gesehen; oder die politischen Eliten als von Nicht-Nativen (Eliten) unterwandert/korrumpiert/beeinflusst. 
Politische und andere Eliten (z.B. Medien, Gerichte) werden als „linkes/wokes Establishment“ kritisiert, das mit Nicht-Nativen unter einer Decke steckt oder diese repräsentiert. 
Nicht-Native können Einwanderer, Asylbewerber und (indigene) ethnische Minderheiten, religiöse Minderheiten, LGBTQI+, Feministen, „Sozialschmarotzer“, Linke/Progressive und ausländische Staaten und internationale Organisationen sein.

\subsubsection*{Links-Rechts - Verständnis von Volk}
\paragraph{Links}
Ansprache insbesondere an ausgegrenzte/ „marginalisierte“ Gruppen wie Einwanderer, LGBTQI, Frauen, Behinderte, Arbeitslose.... 
Betonung der Notwendigkeit der sozialen, wirtschaftlichen und politischen Einbeziehung. 

Das Volk wird als der „Arbeiterklasse“ oder der (sozio-ökonomisch) benachteiligten Mehrheit zugehörig beschrieben.

\paragraph{Rechts}
Der Staat gehört der „nativen“ Gruppe. Die „native“ Gruppe wird häufig durch die Abgrenzung zu Out-Groups definiert, wie
\begin{itemize}
	\item Einwanderer, Asylbewerber und (indigene) ethnische Minderheiten
	\item religiöse Minderheiten
	\item LGBTQI+
	\item Feministen
	\item Sozialhilfeempfänger
	\item Linke/Progressive
	\item Ausländische Staaten und internationale Organisationen
\end{itemize}
Das heißt, dass das native „(deutsche) Volk“ in Abgrenzung zu Migranten, aber auch anderen oben genannten Gruppen genannt wird. Auch beispielsweise Abgrenzung zu Sozialhilfeempfängern („Sozialschmarotzer“) oder Homosexuellen wird als Rechts kategorisiert.

\section{Further Examples}\label{ap_example_sents}
In this appendix, we present additional examples from our corpus. 
These examples are based on a stratified random sample.
We used each sentence's predicted probability to be assigned to one of the four dimensions.
Then, we selected for each dimension three groups consisting of five sentences.
The anchor value was each dimension's threshold, from which we derived the ``edge cases group'' with boundaries of [threshold - .15, threshold + .15].
The other two groups need to be above (or below) those boundaries, representing sentences that are highly likely (or highly unlikely, respectively) to belong to a certain dimension.
Values above each threshold are highlighted in bold. 

\begin{footnotesize}
	\begin{longtable}{p{.1\textwidth}p{.55\textwidth}p{.04\textwidth}p{.04\textwidth}p{.04\textwidth}p{.04\textwidth}}
\caption{Exemplary predictions for a stratified random sample. Predicted values above the thresholds are highlighted in bold. The predicted values are rounded to two decimal points, hence the occasional prediction of 1.0.}\label{tab_stratified_sample}\\
\toprule
Dim & Sentence & Anti-Elite & People-Centric & Host-Left & Host-Right \\
\midrule
\endfirsthead
\toprule
Dim & Sentence & Anti-Elite & People-Centric & Host-Left & Host-Right \\
\midrule
\endhead

\multirow[t]{15}{*}{\textbf{Anti-Elite}} & \parbox[t]{.55\textwidth}{Auch diejenigen, die jetzt schon in der GKV sind und ihre Beiträge alleine zahlen, bekommen diesen Zuschuss bzw. diese Pauschale.} & 0.0 & 0.01 & 0.0 & 0.0 \\\midrule
 & \parbox[t]{.55\textwidth}{Wir stehen dazu, zu helfen.} & 0.0 & 0.01 & 0.0 & 0.0 \\\midrule
 & \parbox[t]{.55\textwidth}{Zu Recht mahnen Sie, Herr Rose, dass es insbesondere Aufgabe der Politik ist, für den notwendigen Zusammenhalt in einer demokratischen Gesellschaft gerade auch durch den Schutz und die gesellschaftliche Gleichstellung von Minderheiten zu sorgen.} & 0.0 & 0.04 & 0.03 & 0.0 \\\midrule
 & \parbox[t]{.55\textwidth}{Deshalb ist das Gesetz, um das es heute geht, notwendig und sinnvoll.} & 0.0 & 0.0 & 0.0 & 0.0 \\\midrule
 & \parbox[t]{.55\textwidth}{Was passiert, wenn die Linke Verantwortung für Wohnungsbaupolitik mitträgt?} & 0.0 & 0.0 & 0.0 & 0.0 \\\midrule
 & \parbox[t]{.55\textwidth}{Sie wollen auch in diesem Bereich die Energiewende abwürgen.} & 0.35 & 0.0 & 0.0 & 0.0 \\\midrule
 & \parbox[t]{.55\textwidth}{Mehr als eine halbe Milliarde Euro Anwaltskosten hat allein der Rechtsstreit der Konzerne mit dem Ministerium verschlungen.} & 0.4 & 0.0 & 0.06 & 0.0 \\\midrule
 & \parbox[t]{.55\textwidth}{Bisher haben Sie alle Ziele gerissen.} & 0.5 & 0.0 & 0.0 & 0.0 \\\midrule
 & \parbox[t]{.55\textwidth}{Vielmehr bekommen wir diesen ganzen Strauß jedes Jahr im Januar um die Ohren gepfeffert.} & \textbf{0.57} & 0.0 & 0.0 & 0.0 \\\midrule
 & \parbox[t]{.55\textwidth}{Ein Wort zur Breitbrandförderung des Bundes.} & \textbf{0.58} & 0.0 & 0.0 & 0.0 \\\midrule
 & \parbox[t]{.55\textwidth}{Es geht hier um eine Krise der Bundesregierung.} & \textbf{0.94} & 0.0 & 0.0 & 0.0 \\\midrule
 & \parbox[t]{.55\textwidth}{Diese bedenkliche Situation, dass das Petitionsrecht durch die Handlungsunfähigkeit der Regierungskoalition monatelang blockiert war, darf sich nicht wiederholen.} & \textbf{0.98} & 0.01 & 0.0 & 0.0 \\\midrule
 & \parbox[t]{.55\textwidth}{Wir sehen an einem ganz einfachen Beispiel, dass Sie unsere Sicherheitsbehörden überhaupt nicht unterstützen, nämlich indem Sie eigene Polizisten anzeigen, weil sie eine rechtmäßige Abschiebung nicht verhindern.} & \textbf{1.0} & 0.0 & 0.01 & 0.01 \\\midrule
 & \parbox[t]{.55\textwidth}{In der Tat will ich ein top finanziertes staatliches Gesundheitssystem und kein staatliches Gesundheitssystem wie in Großbritannien, das wegen Leuten wie Ihnen, die den Neoliberalismus und den Markt anbieten, in dem Zustand ist, in dem es ist.} & \textbf{1.0} & 0.01 & \textbf{0.51} & 0.02 \\\midrule
 & \parbox[t]{.55\textwidth}{Zwar ist es richtig, dass wir nun klassische Stabilisierungspolitik betreiben müssen, um die Wirtschaft wieder auf Kurs zu bringen und die Unternehmen aus ihrer schwierigen Lage herauszuholen, doch die Bundesregierung scheint in ihrer Wirtschaftswerkstatt vor allem ein Werkzeug zu haben, nämlich das Geld deutscher Steuerzahler und künftiger Generationen.} & \textbf{1.0} & 0.02 & 0.33 & 0.03 \\\midrule
\multirow[t]{15}{*}{\textbf{People-Centric}} & \parbox[t]{.55\textwidth}{Auch die sogenannte Opposition in diesem Haus, die Grünen, die Linken und die FDP, macht mit.} & 0.02 & 0.0 & 0.0 & 0.0 \\\midrule
 & \parbox[t]{.55\textwidth}{Seit Jahren häufen sich – und das ist auch bei uns völlig unbestritten – leider solche Fälle von Übertragungen, die den Share Deal vor allem nutzen, um eine Regelung im Grunderwerbsteuergesetz auszunutzen, die viele Jahre zuvor überhaupt kein Problem darstellte.} & 0.02 & 0.0 & 0.01 & 0.0 \\\midrule
 & \parbox[t]{.55\textwidth}{Sie haben jetzt einen Gesetzentwurf des Kabinetts zum Baulandmobilisierungsgesetz aufgegriffen, das wir nachher gleich diskutieren, und gesagt, den Kernpunkt, das Verbot der Umwandlung von Mietwohnungen in Eigentumswohnungen, trägt die Fraktion nicht mit.} & 0.0 & 0.0 & 0.0 & 0.0 \\\midrule
 & \parbox[t]{.55\textwidth}{Damals hatte die Koalition das abgelehnt.} & 0.0 & 0.0 & 0.0 & 0.0 \\\midrule
 & \parbox[t]{.55\textwidth}{Ich werbe um Ihre Unterstützung und fordere die Kollegen der Linken auf, ihre Position zu überdenken und diese Menschen nicht zu vernachlässigen.} & 0.0 & 0.04 & 0.0 & 0.0 \\\midrule
 & \parbox[t]{.55\textwidth}{Große Fleischproduzenten wie Tönnies in Niedersachsen und NRW verdienen sich auf dem Rücken der Arbeitnehmer dumm und dämlich.} & \textbf{1.0} & 0.39 & \textbf{1.0} & 0.01 \\\midrule
 & \parbox[t]{.55\textwidth}{Das trifft ebenso die hohe Zahl der pflegenden Angehörigen, die dringend mehr ambulante professionelle Hilfe benötigen, damit sie ihren Pflegealltag bewältigen können.} & 0.0 & 0.46 & 0.02 & 0.0 \\\midrule
 & \parbox[t]{.55\textwidth}{Diese Menschen arbeiten in der Regel sechs bis sieben Tage in der Woche, und das von morgens bis abends, und am Ende des Monats bleibt kaum etwas übrig.} & 0.0 & \textbf{0.57} & 0.13 & 0.0 \\\midrule
 & \parbox[t]{.55\textwidth}{Schaffen wir ein politisches und gesellschaftliches Klima, in dem wir Menschen unabhängig von ihrer Kultur und Herkunft willkommen heißen.} & 0.0 & \textbf{0.64} & 0.21 & 0.0 \\\midrule
 & \parbox[t]{.55\textwidth}{Weil die Menschen das Vertrauen in die Bundesregierung und ihre Hauruckmaßnahmen verloren haben.} & \textbf{0.99} & \textbf{0.64} & 0.0 & 0.01 \\\midrule
 & \parbox[t]{.55\textwidth}{Gegen Mitarbeiterinnen und Mitarbeiter, gegen Arbeitnehmerinnen und Arbeitnehmer geht das nicht.} & 0.0 & \textbf{0.71} & 0.01 & 0.0 \\\midrule
 & \parbox[t]{.55\textwidth}{„Wir sind das Volk“, das war ein Satz, der Mauern eingerissen hat und zusammengeführt hat, und wir dürfen nicht akzeptieren, dass er missbraucht wird und wieder neue Mauern errichten sowie spalten soll.} & 0.03 & \textbf{0.86} & 0.03 & 0.0 \\\midrule
 & \parbox[t]{.55\textwidth}{Dass sich diese Altenpflegerinnen im Prinzip im Interesse der Patientinnen, im Interesse der Gesellschaft hier starkgemacht haben, muss Ihnen doch zeigen, dass in unserem Rechtssystem Defizite herrschen.} & 0.0 & \textbf{0.87} & 0.01 & 0.0 \\\midrule
 & \parbox[t]{.55\textwidth}{Uns geht es um die Menschen, egal ob sie pflegebedürftig oder Menschen mit Handicap sind.} & 0.0 & \textbf{0.96} & 0.07 & 0.0 \\\midrule
 & \parbox[t]{.55\textwidth}{Dafür sollten wir am heutigen Tage auch mal ein Dankeschön an unsere Bürger sagen.} & 0.0 & \textbf{0.98} & 0.0 & 0.01 \\\midrule
\multirow[t]{15}{*}{\textbf{Host-Left}} & \parbox[t]{.55\textwidth}{Wenn Sie mal von außen auf Deutschland gucken, dann werden Sie feststellen, dass es uns hier deutlich besser geht als den Menschen fast überall woanders.} & 0.0 & 0.2 & 0.0 & 0.01 \\\midrule
 & \parbox[t]{.55\textwidth}{Das liegt objektiv in unserem Interesse.} & 0.0 & 0.01 & 0.0 & 0.0 \\\midrule
 & \parbox[t]{.55\textwidth}{Er hat sein Kabinett vorgestellt.} & 0.0 & 0.0 & 0.0 & 0.0 \\\midrule
 & \parbox[t]{.55\textwidth}{Auch darum geht es bei einer berechtigten Kritik.} & 0.0 & 0.0 & 0.0 & 0.0 \\\midrule
 & \parbox[t]{.55\textwidth}{CETA ist in diesem Bereich, was die Schaffung einer Klageindustrie angeht, verantwortungslos ausgerichtet, im Sinne prinzipieller Verantwortungslosigkeit für die Folgen eigenen wirtschaftlichen Handelns auf Kosten anderer, wo doch in der sozialen Marktwirtschaft Freiheit und Verantwortung Hand in Hand gehen sollten.} & \textbf{0.98} & 0.0 & 0.01 & 0.01 \\\midrule
 & \parbox[t]{.55\textwidth}{Deshalb müssen wir aufpassen, dass wir gerade die vulnerablen, die verletzlichen jungen Frauen, aber auch die Männer in dem Alter von 18 bis 21 Jahren besonders schützen.} & 0.0 & \textbf{0.62} & 0.33 & 0.0 \\\midrule
 & \parbox[t]{.55\textwidth}{Die Arbeitgeber haben den aufrechten Beamten mit einer Denunziationskampagne und mit Mobbing in Krankheit und Frührente getrieben.} & \textbf{0.99} & 0.49 & 0.35 & 0.0 \\\midrule
 & \parbox[t]{.55\textwidth}{Der gesamte Text beschreibt Migrationspolitik ausschließlich aus der Sicht von Migranten.} & 0.0 & \textbf{0.96} & \textbf{0.47} & 0.12 \\\midrule
 & \parbox[t]{.55\textwidth}{Die Proletarier haben nichts … zu verlieren als ihre Ketten.} & 0.0 & \textbf{0.96} & \textbf{0.52} & 0.0 \\\midrule
 & \parbox[t]{.55\textwidth}{Die Stadt gehört den Menschen, die dort wohnen oder dort wohnen wollen, und nicht den Spekulanten.} & 0.01 & \textbf{1.0} & \textbf{0.56} & 0.04 \\\midrule
 & \parbox[t]{.55\textwidth}{Deswegen muss ganz schnell etwas passieren, damit die Flüchtlinge auf einen entsprechenden bezahlbaren Wohnraum verteilt werden und auch zu Freunden und Familienangehörigen gehen können.} & 0.0 & \textbf{0.87} & \textbf{0.59} & 0.01 \\\midrule
 & \parbox[t]{.55\textwidth}{Wir müssen auf die setzen, die nicht das große Glück reicher Eltern haben und auf die Unterstützung des Staates angewiesen sind.} & 0.0 & \textbf{0.97} & \textbf{0.89} & 0.0 \\\midrule
 & \parbox[t]{.55\textwidth}{Stoppen wir Elektromobilität zum Schutz von rohstoffreichen Entwicklungsländern, zum Schutz ihrer Menschen, gerade ihrer Kinder.} & 0.06 & \textbf{0.98} & \textbf{0.92} & 0.0 \\\midrule
 & \parbox[t]{.55\textwidth}{Dass Sie sich aus Angst vor einer dicken, fetten Schlagzeile in einer großen Zeitung genau in diesem Moment auch als Finanzminister wegducken, zeigt, dass Sie offensichtlich aus Warburg und Wirecard nicht wirklich etwas gelernt haben.} & \textbf{1.0} & 0.01 & \textbf{0.97} & 0.0 \\\midrule
 & \parbox[t]{.55\textwidth}{Wir müssen unseren Blick aber genauso auf die Situation der Menschen richten, die schon lange ohne Arbeit sind.} & 0.0 & \textbf{1.0} & \textbf{0.98} & 0.01 \\\midrule
\multirow[t]{15}{*}{\textbf{Host-Right}} & \parbox[t]{.55\textwidth}{Wenn tatsächlich nur 5 Prozent der Optionspflichtigen von der neuen Regelung betroffen sind, warum nehmen wir dann die restlichen 95 Prozent in Haftung?} & 0.0 & 0.0 & 0.0 & 0.0 \\\midrule
 & \parbox[t]{.55\textwidth}{Deswegen muss der 52-GW-Deckel weg, und die von Minister Altmaier versprochene Südquote muss dringend her.} & 0.0 & 0.0 & 0.0 & 0.0 \\\midrule
 & \parbox[t]{.55\textwidth}{Das sind fünf Fakten einer wissenschaftlichen Studie, die gegen die Erforderlichkeit sprechen.} & 0.0 & 0.0 & 0.0 & 0.0 \\\midrule
 & \parbox[t]{.55\textwidth}{In der Sache sind wir uns aber weitgehend einig: Die Weichen werden in Richtung mehr Verkehr auf der Schiene gestellt, und in Lärmschutzbelangen kommen wir den Menschen im Rheintal, aber auch im Rest der Republik deutlich entgegen.} & 0.0 & 0.01 & 0.0 & 0.0 \\\midrule
 & \parbox[t]{.55\textwidth}{Das geht auch aus dem Europol-Bericht aus dem Jahr 2018 hervor, der die Verbindung zwischen organisierter Kriminalität und der Hisbollah offenlegt, beispielsweise aus Einkünften aus Geldwäsche zum Mittel der Terrorfinanzierung.} & 0.01 & 0.0 & 0.0 & 0.0 \\\midrule
 & \parbox[t]{.55\textwidth}{Während der grünen Regierungszeit gab es im Bereich der Forschung und Bildung schlichtweg nur finanzielle Stagnation, nicht mehr und nicht weniger.} & \textbf{0.98} & 0.0 & 0.0 & 0.23 \\\midrule
 & \parbox[t]{.55\textwidth}{Deswegen sind der deutsche Diesel, die deutsche Umwelttechnik wie auch die saubersten Kohle- und Gaskraftwerke der Welt – alles „made in Germany“ – echter Umweltschutz für echte Entwicklungspolitik, gerade in Afrika.} & 0.01 & \textbf{0.51} & 0.0 & 0.24 \\\midrule
 & \parbox[t]{.55\textwidth}{Wenn es nach den Vorstellungen von Grünen oder Linken geht, soll es nämlich bald vorbei sein mit brutto für netto.} & 0.25 & 0.01 & 0.0 & 0.24 \\\midrule
 & \parbox[t]{.55\textwidth}{Der nationale Bildungsbericht hält für die Zuzugsjahre von 2014 bis 2016 erschreckende Zahlen bereit: 69 Prozent der Migranten ab 15 Jahren aus diesen Ländern haben keinen Berufsabschluss und natürlich auch keinen höheren Bildungsabschluss.} & 0.02 & 0.06 & 0.01 & 0.26 \\\midrule
 & \parbox[t]{.55\textwidth}{Auch die Zahl derer – jetzt komme ich wieder zu unserem Lieblingsthema –, die seit 2015 ungebremst nach Deutschland einströmen, verursacht hohe Kosten für medizinische Leistungen.} & 0.01 & 0.14 & 0.0 & \textbf{0.48} \\\midrule
 & \parbox[t]{.55\textwidth}{Es ist außerdem eine Frechheit, dass die Linksfraktion steuerliche Vergünstigungen für ihr politisches Umfeld möchte, obwohl sie sich milliardenschweres SED-Vermögen unter den Nagel gerissen hat.} & \textbf{1.0} & 0.01 & \textbf{0.92} & \textbf{0.91} \\\midrule
 & \parbox[t]{.55\textwidth}{Es sind nicht Migranten das Problem, sondern die Parallelgesellschaften, die mit der deutschen Sprache und deutschen Tugenden nichts anfangen können.} & 0.02 & \textbf{0.94} & 0.05 & \textbf{0.95} \\\midrule
 & \parbox[t]{.55\textwidth}{Eine europäische Armee, ein EU-Finanzminister, eine gemeinsame Bankenhaftung, ein gemeinsamer Währungsfonds, eine gemeinsame Arbeitslosenversicherung – der Fantasie der Transferleistungen von Nordeuropa an den Club Med, der von Frankreich orchestriert wird, sind scheinbar keine Grenzen gesetzt.} & \textbf{1.0} & 0.0 & 0.06 & \textbf{0.96} \\\midrule
 & \parbox[t]{.55\textwidth}{Stellen Sie also diese Signale ab, und schützen Sie die deutschen Grenzen gegen illegale Migranten!} & \textbf{0.65} & 0.2 & 0.0 & \textbf{0.98} \\\midrule
 & \parbox[t]{.55\textwidth}{Also, liebe Kolleginnen und Kollegen der Grünen, Sie haben die Wahl: Weitermachen mit der grünen Anti-Massentierhaltungspropaganda, die jedes konstruktive Gespräch mit Bürgern zum Thema Stallneubau unmöglich macht, oder ein ehrliches Agieren Ihrerseits im Sinne unserer Umwelt, und zwar gemeinsam mit unseren Landwirten und nicht gegen sie.} & \textbf{1.0} & \textbf{0.63} & 0.02 & \textbf{0.99} \\
\bottomrule
\end{longtable}
\end{footnotesize}

\section{Original Figures}\label{ap_orig_figures}
\FloatBarrier
The axes in the figures within the main text were normalized to their respective maximum values.
For the sake of transparency, we present here the figures with the unstandardized axes once again for interested readers.

\begin{figure*}
	\centering
	\includegraphics[width=\linewidth]{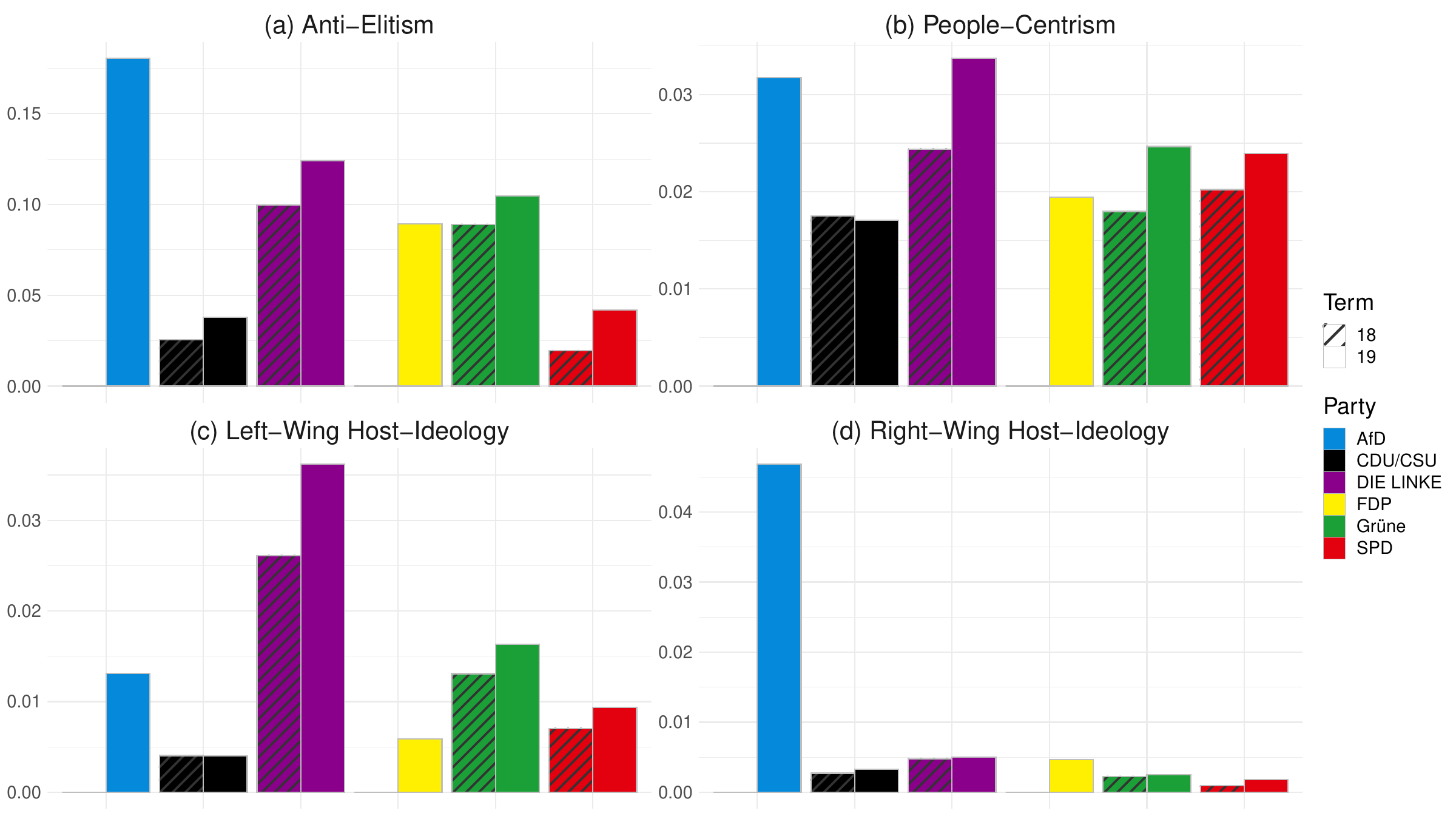}
	\caption{Populist dimensions in speeches of the 18th and 19th legislative period of the Bundestag, by party. This figure shows the same values as Figure \ref{fig_all_dimensions} in the original paper, but with unstandardized axes; hence, the different scales.}
\end{figure*}

\begin{figure*}
	\centering
	\includegraphics[width=\linewidth]{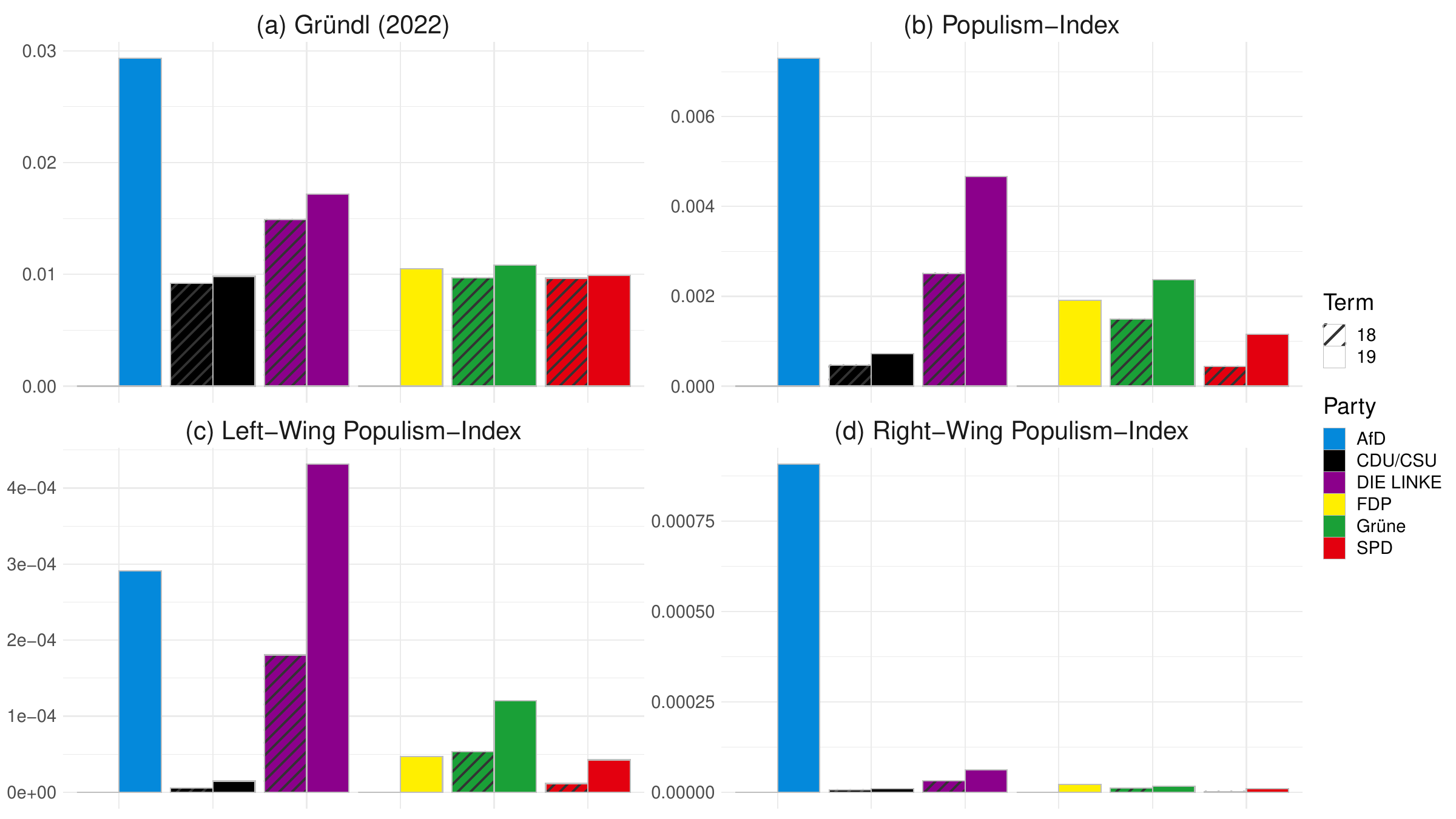}
	\caption{Populism in speeches of the 18th and 19th legislative period of the Bundestag, by party. This figure shows the same values as Figure \ref{fig_multi_index} in the original paper but with unstandardized axes; hence, the different scales.}
\end{figure*}

\FloatBarrier

\section{External Validation}\label{apextval}
This appendix presents the 17 prototypical populist statements extracted from other articles that we used as out-of-sample test.\footnote{Idioms have been translated with their German equivalents (e.g. ``Washington'' to ``Berlin'', ``SNCF'' to ``Bahn'',  ``the American people'' to ``Bürger''). }
We relied on the following studies to extract sentences:

The analysis in \textcite{schurmannYellingSidelinesHow2022} was conducted as a manual content analysis of 3500 Facebook posts from prominent German politicians and parties. 
While the article differentiates between various dimensions of populism throughout, this distinction is not evident in the supplementary material for the examples. 
Therefore, in this study, a post is considered a positive match if it exhibits at least one of the two key dimensions of populism. 
All posts from \textcite{schurmannYellingSidelinesHow2022} resonate with at least one of our identified dimensions.

In a semi-automated content analysis, \textcite{ernstExtremePartiesPopulism2017} examined 1400 Facebook posts from six Western European countries to study populist communication strategies.
The examples provided in their supplementary materials comprise six posts that align with our categorization into dimensions.
Among these six posts, we identified an exact match for the specified dimensions in five of them. 
Our analysis did not recognize one post from the dimension of people-centrism.

\textcite{bonikowskiPoliticsUsualMeasuring2022} conducted a study on election speeches in the USA and provided two examples during the explanation of their theoretical concept in the article to illustrate populism.
We identified at least one of our two dimensions for both examples. 

\textcite{daiWhenPoliticiansUse2022} also investigated election speeches in the USA and presented two examples. 
In both cases, our two dimensions of populism were also applicable and evident.

\begin{footnotesize}
	\begin{longtable}{p{.1\textwidth}p{.07\textwidth}p{.45\textwidth}p{.04\textwidth}p{.04\textwidth}p{.04\textwidth}p{.04\textwidth}}
\caption{Predictions on populist sentences from other studies. Predicted values above the threshold are highlighted in bold. The predicted values are rounded to two decimal points, hence the occasional prediction of 1.0.}\label{tab_external_sent_val}\\
\toprule
 Source & Label & Text & Anti-Elite & People-Centric & Host-Left & Host-Right \\
\midrule
\endfirsthead
\toprule
 Source & Label & Text & Anti-Elite & People-Centric & Host-Left & Host-Right \\
\midrule
\endhead

\multirow[t]{7}{*}{\parbox[t]{.1\textwidth}{Schürmann and Gründl 2022}} & \multirow[t]{7}{*}{\parbox[t]{.07\textwidth}{\textbf{populism}}} & \parbox[t]{.45\textwidth}{"Die da oben" bestimmen über "uns hier unten"? Dieses Gefühl der Ohnmacht vieler Bürger will die AfD aufheben. Einer der mir wichtigsten Punkte in unserem Wahlprogramm ist deshalb dieser: Wir wollen dem Volk das Recht geben, den Abgeordneten auf die Finger zu schauen und vom Parlament beschlossene Gesetze zu ändern oder abzulehnen. Das Volk soll auch die Möglichkeit erhalten, eigene Gesetzesinitiativen einzubringen und per Volksabstimmung zu beschließen. \\\textit{The one up there” controls “us down here”? This feeling of disempowerment for many citizens is what the AfD wants to overcome. One of the most important points in our election program is therefore this: "We want to give the people the right to watch the MPs and to change or reject laws passed by parliament. The people should also have the opportunity to introduce their own legislative initiatives and pass them by referendum.}} & 0.01 & \textbf{1.0} & 0.02 & 0.01 \\\midrule
 &  & \parbox[t]{.45\textwidth}{Ich denke, für Familien und Menschen mit kleinen und mittleren Einkommen müsste noch mehr getan werden: Angesichts der Rücklagen die Sozialabgaben senken, den Soli schrittweise abschaffen, die kalte Progression zurückführen, Kindergeld erhöhen, Baukindergeld umsetzen und vieles mehr. Wir müssen die Normalbürger in der Mitte entlasten.\\\textit{I think more should be done for families and people with small and medium salaries: In light of our reserves, [we must] lower social welfare contributions, gradually abolish the Soli [solidarity tax for former DDR states], reduce the cold progression, increase child benefits, implement Baukindergeld [state subsidy for families buying a house] and much more. We need to unburden the ordinary people in the middle class."}} & 0.0 & \textbf{0.94} & 0.05 & 0.0 \\\midrule
 &  & \parbox[t]{.45\textwidth}{Integration kann nicht bedeuten, dass sich die einheimische Bevölkerung und die Zuwanderer auf halbem Weg treffen und daraus eine neue Kultur entsteht. Wir brauchen bei der Integration in Deutschland einen klaren Kompass: unsere Leitkultur. Eine Mehrheit der Deutschen sieht das genauso!\\\textit{Integration does not mean that the natives and immigrants meet halfway and a new culture emerges from this. We need a clear direction for integration in Germany: our Leitkultur. A majority of Germans thinks like us!}} & 0.0 & \textbf{0.98} & 0.03 & \textbf{0.43} \\\midrule
 &  & \parbox[t]{.45\textwidth}{Die CSU hat zweieinhalb Jahre lang nichts getan und tritt jetzt, vor der bayerischen Landtagswahl, hart auf. Die Bürgerinnen und Bürger in Bayern sollten es sich nicht bieten lassen, für einen Landtagswahlkampf so instrumentalisiert zu werden.\\\textit{The CSU has done nothing for two and a half years, an now ahead of the Bavarian state election, they are getting tough. The citizens of Bavaria should not tolerate to get instrumentalized for a state election campaign like this.}} & \textbf{0.99} & \textbf{0.58} & 0.0 & 0.01 \\\midrule
 &  & \parbox[t]{.45\textwidth}{CDU/CSU und SPD stellen kurzfristige Konzerninteressen über den Gesundheitsschutz der Menschen. Mit dieser Kumpanei muss endlich Schluss sein!\\\textit{The CDU/CSU and SPD are putting short-term interests of big companies over the protection of people's health. This cronyism must come to an end!}} & \textbf{1.0} & 0.45 & \textbf{1.0} & 0.01 \\\midrule
 &  & \parbox[t]{.45\textwidth}{Warum Konzerne an Parteien spenden, lässt sich gerade beobachten. Die Politiker der Großen Koalition verhalten sich wie Lobbyisten der Autoindustrie. Beschwichtigung und Aussitzen, statt Aufklärung und Verbraucherschutz. DIE LINKE hingegen ist nicht käuflich! Bei den anderen Parteien zahlen Unternehmen und Wirtschaftsverbände – bei uns zählst Du! \\\textit{Why large concerns donate to political parties can be observed right now. The politicians of the grand coalition act like lobbyists for the car industry. Appeasement and waiting, instead of information and consumer protection. Instead, DIE LINKE is not for sale! With the other parties, companies and trade associations pay - with us, you matter! }} & \textbf{1.0} & 0.2 & \textbf{1.0} & 0.04 \\\midrule
 &  & \parbox[t]{.45\textwidth}{Ich bin überzeugt, dass die Verunsicherung in der Bevölkerung ganz viel mit dem Wandel in der Arbeitswelt zu tun hat. Wir werden diesen Wandel gestalten und zwar gemeinsam mit den Menschen. Die SPD braucht bei der der Zukunft der Arbeit eine klarere Profilierung.\\\textit{I am convinced that the uncertainty among the population has a lot to do with the transformation of work. We will shape this transformation, and we will do so together with the people. The SPD needs a clearer profile when it comes to the future of labour.}} & 0.0 & \textbf{0.59} & 0.0 & 0.0 \\\midrule
\multirow[t]{2}{*}{\parbox[t]{.1\textwidth}{Bonikowski et al. 2022}} & \multirow[t]{2}{*}{\parbox[t]{.07\textwidth}{\textbf{populism}}} & \parbox[t]{.45\textwidth}{Schließlich müssen die Bürger darauf vertrauen können, dass ihre Regierung sich um uns alle kümmert - und nicht um die Sonderinteressen, die seit acht Jahren die Agenda in Berlin bestimmen, und um die Lobbyisten, die die Kampagne leiten. Ich habe meine Karriere damit verbracht, es mit Lobbyisten und deren Geld aufzunehmen, und ich habe gewonnen.\\\textit{Finally, the American people must be able to trust that their government is looking out for all of us—not the special interests that have set the agenda in Washington for eight years, and the lobbyists who run John McCain’s campaign. I’ve spent my career taking on lobbyists and their money, and I’ve won}} & \textbf{1.0} & \textbf{0.73} & \textbf{1.0} & 0.01 \\\midrule
 &  & \parbox[t]{.45\textwidth}{Es wird ein Sieg für das Volk sein. Ein Sieg für den einfachen Bürger, dessen Stimme bisher nicht gehört wurde. Es wird ein Sieg für die Wähler sein, nicht für die Experten, nicht für die Journalisten, nicht für die Lobbyisten, nicht für die globalen Sonderinteressen, die die Kampagne meines Gegners finanzieren.\\\textit{It’s going to be a victory for the people. A victory for the everyday citizen whose voice hasn’t been heard. It will be a win for the voters, not the pundits, not the journalists, not the lobbyists, not the global special interests funding my opponent’s campaign.}} & 0.34 & \textbf{1.0} & 0.29 & 0.04 \\\midrule
\multirow[t]{2}{*}{\parbox[t]{.1\textwidth}{Dai and Kustov 2022}} & \multirow[t]{2}{*}{\parbox[t]{.07\textwidth}{\textbf{populism}}} & \parbox[t]{.45\textwidth}{Der Wandel wird kommen. All die Leute, die das System zu ihrem persönlichen Vorteil manipuliert haben, versuchen, unsere Veränderungskampagne zu stoppen, weil sie wissen, dass ihr Geldsegen am letzten Halt angekommen ist. Jetzt bist du dran. Dies ist Ihre Zeit ... Wir kämpfen für alle Bürger, ... die von diesem korrupten System im Stich gelassen wurden. Wir kämpfen für alle, die keine Stimme haben. Hillary Clinton ist die Kandidatin der Vergangenheit. Unsere ist die Kampagne der Zukunft. In dieser Zukunft werden wir eine neue Handelspolitik verfolgen, bei der die Arbeitnehmer an erster Stelle stehen - und bei der die Arbeitsplätze in unserem Land erhalten bleiben ... Die Zeit der wirtschaftlichen Kapitulation ist vorbei.\\\textit{Change is coming. All the people who’ve rigged the system for their own personal benefit are trying to stop our change campaign because they know that their gravy train has reached its last stop. It’s your turn now. This is your time ... We are fighting for all Americans ... who’ve been failed by this corrupt system. We’re fighting for everyone who doesn’t have a voice. Hillary Clinton is the candidate of the past. Ours is the campaign of the future. In this future, we are going to pursue new trade policies that put American workers first – and that keep jobs in our country ... The era of economic surrender is over.}} & \textbf{1.0} & \textbf{0.89} & \textbf{0.94} & 0.0 \\\midrule
 &  & \parbox[t]{.45\textwidth}{Drittens habe ich gesagt, dass wir die Finanzindustrie nicht einfach retten können und werden, ohne den Millionen von unschuldigen Hausbesitzern zu helfen, die darum kämpfen, in ihren Häusern zu bleiben. ... Ich habe gesagt, dass ich nicht zulassen werde Ich habe gesagt, dass ich nicht zulassen werde, dass dieser Plan zu einem Wohlfahrtsprogramm für die Führungskräfte der Finnazindustrie wird, deren Gier und Verantwortungslosigkeit uns in diesen Schlamassel gebracht haben. ... Wir brauchen nicht nur einen Plan für Banker und Investoren, wir brauchen einen Plan für Autoarbeiter, Lehrer und Kleinunternehmer. ... Das bedeutet, dass wir es mit den Lobbyisten und Sonderinteressen in Berlin aufnehmen müssen. Das bedeutet, gegen die Gier und Korruption in der Finanzindustrie vorzugehen ... Es ist an der Zeit, Berlin zu reformieren.\\\textit{Third, I said that we cannot and will not simply bailout Wall Street without helping the millions of innocent homeowners who are struggling to stay in their homes. ... I said that I would not allow this plan to become a welfare program for the Wall Street executives whose greed and irresponsibility got us into this mess. ... We don’t just need a plan for bankers and investors, we need a plan for autoworkers and teachers and small business owners. ... That means taking on the lobbyists and special interests in Washington. That means taking on the greed and corruption on Wall Street ... It is time to reform Washington.}} & \textbf{0.99} & \textbf{0.76} & \textbf{1.0} & 0.03 \\\midrule
\multirow[t]{6}{*}{\parbox[t]{.1\textwidth}{Ernst et al. 2017}} & \multirow[t]{2}{*}{\parbox[t]{.07\textwidth}{\textbf{populism}}} & \parbox[t]{.45\textwidth}{Erst die anderen, dann wir: Für illegale Einwanderer ist die Zugfahrt kostenlos! Danke, Bahn!\\\textit{The others before us: Trains for the illegal are for free! Thank you SNCF!}} & \textbf{0.99} & 0.32 & 0.04 & \textbf{0.99} \\\midrule
 &  & \parbox[t]{.45\textwidth}{Volle Unterstützung für unsere Bauern, die heute erneut von der EU und der Regierung gedemütigt wurden. Es ist Zeit für eine deutsche Agrarpolitik!\\\textit{Unlimited support for our peasant once more humiliated by the EU and our government today. Long live the French agrarian politic.}} & \textbf{0.99} & \textbf{0.94} & 0.1 & 0.01 \\\midrule
 & \multirow[t]{2}{*}{\parbox[t]{.07\textwidth}{\textbf{anti-elitism}}} & \parbox[t]{.45\textwidth}{Die Regierung weiß längst nicht mehr, wo sie diskriminiert und wenn ja warum.\\\textit{The Government no longer knows, where it discriminates and even if, not why}} & \textbf{1.0} & 0.0 & 0.01 & 0.01 \\\midrule
 &  & \parbox[t]{.45\textwidth}{Der unverantwortliche Umgang der EU mit der Migrantenkrise verursacht Chaos und ist ein weiteres Beispiel dafür, warum wir austreten müssen.\\\textit{The EU's irresponsible approach to the migrant crisis is causing chaos and is another example of why we must leave.}} & \textbf{1.0} & 0.04 & 0.01 & \textbf{0.95} \\\midrule
 & \multirow[t]{2}{*}{\parbox[t]{.07\textwidth}{\textbf{people-centrism}}} & \parbox[t]{.45\textwidth}{Schriftsteller Peter Stamm unterstützt mich, weil ich "ein Volksvertreter und kein Wirtschaftsvertreter" bin. \\\textit{Author Peter Stamm supports me because I am "a representative of the people and not an economic representative".}} & 0.0 & 0.0 & 0.0 & 0.0 \\\midrule
 &  & \parbox[t]{.45\textwidth}{Da ich meine Kampagne selbst finanziere, werde ich nicht von meinen Spendern, besonderen Interessen oder Lobbyisten kontrolliert. Ich arbeite nur für die Menschen in Deutschland!\\\textit{By self-funding my campaign, I am not controlled by my donors, special interests or lobbyists. I am only working for the people of the U.S.!}} & 0.01 & \textbf{1.0} & 0.03 & 0.02 \\
\bottomrule
\end{longtable}
\end{footnotesize}

\end{document}